%% file: main.tex
\documentclass{sjtudenglab}

\usepackage{amsmath}
\usepackage{amsfonts}
\usepackage{amssymb}
\usepackage{amsthm}
\usepackage{mathtools}
\usepackage{enumerate}
\usepackage{enumitem}
\usepackage{algorithm}
\usepackage{algpseudocode}
\usepackage{newtxtt}
\usepackage{diagbox}
\usepackage{colortbl}
\usepackage{xspace}
\usepackage{wrapfig}
\usepackage{adjustbox}
\usepackage{tabularx}
\usepackage{booktabs}
\usepackage{multirow}
\usepackage{makecell}
\usepackage{tikz}
\usepackage{dsfont}
\usepackage{xfakebold}
\usepackage{natbib}
\usepackage{siunitx}
\usepackage{thmtools}
\usepackage{wrapfig}

\usepackage[most]{tcolorbox}
\usepackage{enumitem}
\usepackage{fontawesome5}
\usepackage{xcolor}

\definecolor{mbdteal}{RGB}{0,105,110}
\definecolor{mbdtealbg}{RGB}{244,250,250}

\tcbset{
    contributionbox/.style={
        colback=mbdtealbg,
        colframe=mbdteal,
        coltitle=white,
        colbacktitle=mbdteal,
        boxrule=0.5pt,
        arc=3pt,
        left=6pt,
        right=6pt,
        top=6pt,
        bottom=6pt,
        fonttitle=\bfseries,
        titlerule=0pt
    }
}

\definecolor{reqblue}{RGB}{54,96,146}
\definecolor{reqbg}{RGB}{246,249,253}

\tcbset{
    requirementbox/.style={
        colback=reqbg,
        colframe=reqblue,
        coltitle=white,
        colbacktitle=reqblue,
        boxrule=0.5pt,
        arc=3pt,
        left=6pt,
        right=6pt,
        top=5pt,
        bottom=5pt,
        fonttitle=\bfseries,
        titlerule=0pt
    }
}

\usepackage[dvipsnames,table]{xcolor}

\usetikzlibrary{positioning, calc}
\usetikzlibrary{decorations.pathmorphing}

\usepackage{tcolorbox}
\tcbuselibrary{minted}
\setminted[python]{
  linenos,
  breaklines,
  fontsize=\footnotesize,
  xleftmargin=2em
}

\usepackage{silence}
\usepackage{hyperref}
\usepackage{cleveref}

\makeatletter
\AtBeginDocument{
  \urlstyle{sf}
  
}
\makeatother

\numberwithin{equation}{section}
\floatstyle{plain}
\newfloat{algorithmgroup}{tbp}{loag}
\floatname{algorithmgroup}{Algorithm Group}

\definecolor{light}{RGB}{125, 125, 125}
\definecolor{cverbbg}{gray}{0.90}
\definecolor{textgray}{HTML}{6E6E73}
\definecolor{trainingfreegreen}{RGB}{238,248,238}
\definecolor{multibdblue}{RGB}{235,245,255}
\definecolor{commentcolor}{RGB}{77,99,135}

\newtcolorbox[auto counter]{pbox}[2][]{
  colback=white,
  title=Code~\thetcbcounter: #2,
  #1,
  fonttitle=\sffamily,
  fontupper=\sffamily,
  arc=2pt,
  colframe=bgcolor,
  coltitle=fgcolor,
  colbacktitle=bgcolor,
  toptitle=0.25cm,
  bottomtitle=0.125cm
}

\makeatletter
\newcommand\applefootnote[1]{%
  \begingroup
  \renewcommand\thefootnote{}%
  \renewcommand\@makefntext[1]{\noindent##1}%
  \footnote{#1}%
  \addtocounter{footnote}{-1}%
  \endgroup
}
\makeatother

\crefname{tcb@cnt@pbox}{code}{code}
\Crefname{tcb@cnt@pbox}{Code}{Code}
\crefname{assumption}{assumption}{assumption}
\Crefname{assumption}{Assumption}{Assumptions}

\title{Multi-Block Diffusion Language Models}

\author[1]{Yijie Jin}
\author[2]{Jiajun Xu}
\author[1]{Yuxuan Liu}
\author[1]{Chenkai Xu}
\author[3]{Yi Tu}
\author[3]{Jiajun Li}
\author[3]{Dandan Tu}
\author[3]{Xiaohui Yan}
\author[1]{Kai Yu}
\author[1]{Pengfei Liu}
\author[1,\dagger]{Zhijie Deng}

\affiliation[1]{Shanghai Jiao Tong University}
\affiliation[2]{Xi'an Jiao Tong University}
\affiliation[3]{Huawei}

\abstract{
\input{sections/00_abstract}
}

\metadata[Project Page]{\url{https://sjtu-deng-lab.github.io/mbd-lms}} 

\metadata[Correspondence]{\sffamily Zhijie Deng: \href{mailto:zhijied@sjtu.edu.cn}{zhijied@sjtu.edu.cn}}

\metadata[Contributions]{\sffamily$^\dagger$ Corresponding author.}

\date{\sffamily\today}

\AtBeginDocument{ 
  \fancyheadoffset[R]{-0.3cm}
  \fancyhead[R]{\raisebox{9pt}{\includegraphics[height=35pt]{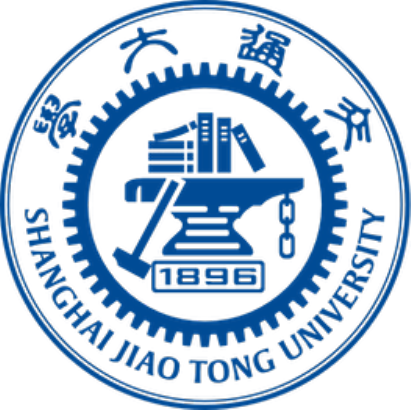}}}
}

\begin{document}

\maketitle

\input{sections/01_introduction}
\input{sections/02_preliminary}
\input{sections/04_methodology}
\input{sections/05_experiments}

\input{sections/06_related_work}
\input{sections/07_conclusion}

\setlength{\bibsep}{5pt}
\bibliography{custom}
\bibliographystyle{plainnat}

\newpage
\appendix
\input{sections/08_appendix_proof}
\input{sections/09_appendix_training_implementation}
\input{sections/10_appendix_system_design_detail}
\input{sections/11_appendix_experiments}

\end{document}

%% file: sections/00_abstract.tex

Block Diffusion Language Models (BD-LMs) improve diffusion-based text generation with KV caching and flexible-length generation. 
A natural next step is to extend them from Single-Block Diffusion (SingleBD) to Multi-Block Diffusion (MultiBD), where a \textit{running-set} of consecutive blocks is decoded concurrently for inter-block parallelism. 
However, existing BD-LMs are mostly trained under teacher forcing, where the model observes only one noisy block conditioned on a clean prefix. 
While the recent diffusion forcing strategy introduces visibility among multiple noisy blocks, its training states still differ from MultiBD inference, where decoding operates on a bounded \textit{running-set} with heterogeneous slot-wise noise patterns. 
To bridge this gap, we propose \textit{Multi-Block Diffusion Language Models} (MBD-LMs), obtained by post-training BD-LMs with \textit{Multi-block Teacher Forcing} (MultiTF).
MultiTF integrates teacher forcing and diffusion forcing by training on bounded \textit{noise-groups} conditioned on clean prefixes, with randomized \textit{noise-schedulers} that better match MultiBD inference states. 
To make MultiBD practically executable, we further introduce an optimized decoding algorithm based on the \textit{Block Buffer} mechanism that preserves prefix-cache reuse, keeps input shapes static, and translates increased decoding parallelism into wall-clock acceleration. 
Empirically, MBD-LLaDA2-Mini increases average Tokens Per Forward pass (TPF) from 3.47 to \textbf{6.19} and improves average accuracy from 79.95\% to \textbf{81.03\%}; when combined with DMax, MBD-LLaDA2-Mini-DMax reaches an average TPF of \textbf{9.34} with only a 1.02\% accuracy drop on math and code benchmarks.

%% file: sections/01_introduction.tex
\section{Introduction}

\begin{figure}[t]
  \centering
  \includegraphics[width=0.9\columnwidth]{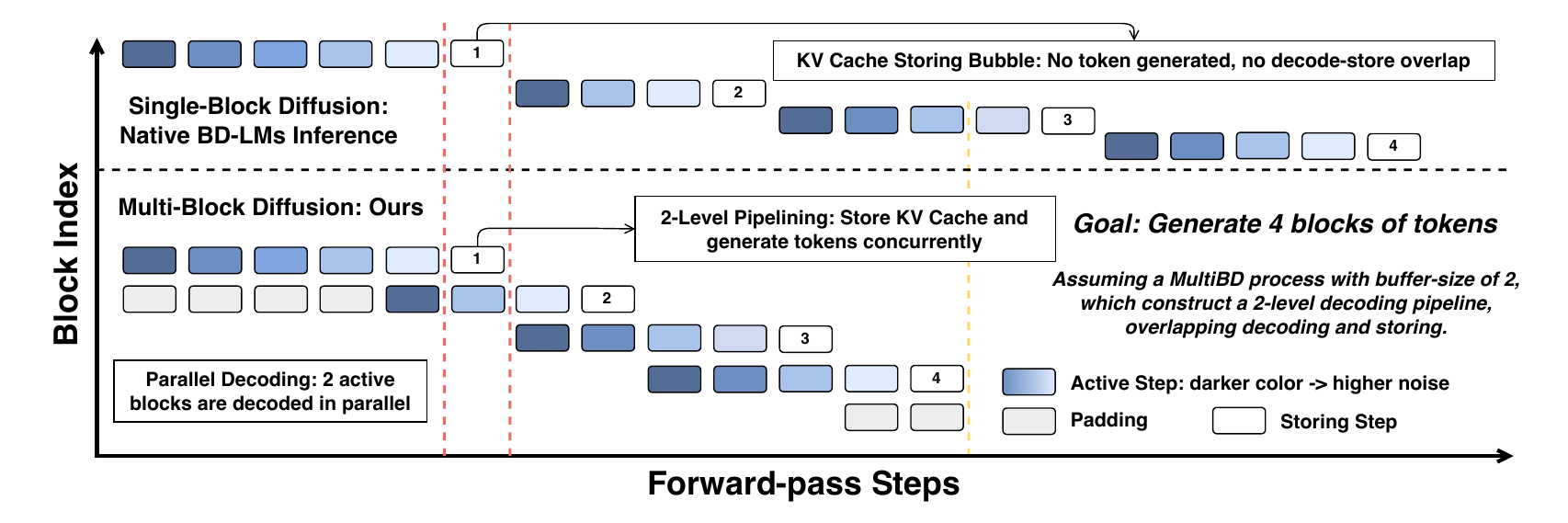}
  \caption{
    SingleBD decodes blocks sequentially and creates KV cache storing bubbles.
    In contrast, MultiBD overlaps future-block refinement with KV cache storing of completed blocks, and enables inter-block parallelism.
    }
  \label{fig:single_vs_multiblock_pipeline}
\end{figure}

Diffusion Language Models (DLMs) have emerged as a promising alternative to autoregressive language models by enabling native parallel decoding~\citep{sahoo2024mdlm,nie2025llada}. 
However, fully bidirectional DLMs struggle to serve efficiently because they lack support for KV caching and dynamic-length generation.

Recent Block Diffusion Language Models (BD-LMs) have become a representative DLM paradigm for efficient generation, addressing the above limitations through block-causal generation~\citep{arriola2025block,bie2025llada2,cheng2025sdar}.
Most BD-LMs trained under Teacher Forcing (TF) naturally support \textit{Single-Block Diffusion} (SingleBD): at each forward pass, the model decodes one noisy block while preceding blocks are already clean and cached, enabling KV caching and intra-block parallelism. 
However, blocks themselves are still processed sequentially.
As shown in Figure~\ref{fig:single_vs_multiblock_pipeline}, SingleBD must finish decoding a block and storing its KV cache before later blocks can proceed, creating storing bubbles and locking inter-block parallelism.


The Discrete Diffusion Forcing (D2F)~\citep{wang2025d2f} strategy introduces the visibility of multiple noisy blocks to BD-LMs.
Conditioned on a clean prefix, it corrupts suffix blocks with monotonic increasing noise ratios during training.
Consequently, D2F obtains \textit{Multi-Block Diffusion} (MultiBD) capability, as shown in Figure~\ref{fig:single_vs_multiblock_pipeline}, enabling \underline{decode-store overlap} and \underline{inter-block parallelism}.
However, a \underline{train--inference mismatch} problem remains. 
Specifically, it is not possible to process the entire noisy suffix as one \textit{running-set} in a single forward pass, from both the perspectives of efficiency and empirical efficacy~\citep{lu2026adablock}.
For the naive MultiBD introduced by D2F, the expected \textit{running-set} size is often around two, and adjacent slots exhibit large noise-ratio gaps.  
This suggests that reliable MultiBD requires training states that match both the bounded \textit{running-set} size and the heterogeneous slot-wise noise patterns observed during inference.


To this end, we formulate \textit{Multi-Block Diffusion Language Models} (MBD-LMs), a unified view of existing BD-LMs.
This view covers both TF-trained BD-LMs and D2F-trained BD-LMs as extreme cases, while identifying practical MultiBD as the bounded intermediate regime for reliable and efficient inference.

We introduce \textit{Multi-block Teacher Forcing} (MultiTF), a post-training method that turns BD-LMs into MBD-LMs. 
MultiTF extends TF by concatenating the clean prefix with a bounded group of consecutive noisy blocks, where noisy blocks can attend to each other under a \textit{Group-Aware Dual-Stream Mask}. 
It applies a more aggressive and randomized \textit{noise-scheduler} within each \textit{noise-group} to simulate the heterogeneous slot-wise noise patterns observed during inference.  
During training, blocks are partitioned into groups with varying sizes to cover possible \textit{running-set} sizes and group-relative positions.

We further propose an optimized inference pipeline for MultiBD.
MultiBD relies on a dynamic \textit{running-set} for decoding, which is unfriendly to \underline{CUDA Graph capture and replay}.
To address this, we introduce the \textit{Block Buffer} mechanism, which maintains a fixed number of block slots. 
Future blocks enter the \textit{Block Buffer} by activating existing idle slots rather than extending the physical input, while completed front blocks leave after being committed to the KV cache.
This design keeps the \underline{input shape static}, preserves KV caching and \underline{prefix caching}, and translates the increased TPF into practical wall-clock speedup.



Experiments on math and code benchmarks show that MBD-LMs improve decoding parallelism while preserving generation quality.
Compared with LLaDA2-Mini~\citep{bie2025llada2}, \textbf{MBD-LLaDA2-Mini} increases the average TPF from 3.47 to \textbf{6.19} (+78.4\%) and improves the average accuracy from 79.95\% to \textbf{81.03\%}. 
When combined with DMax~\citep{chen2026dmax}, \textbf{MBD-LLaDA2-Mini-DMax} further reaches an average TPF of \textbf{9.34} (+47.1\% over LLaDA2-Mini-DMax under SingleBD) with only a 1.02 percentage-point accuracy drop. 
Using our inference engine, MBD-LLaDA2-Mini-DMax achieves 951.41 TPS on average, compared with 781.50 TPS for LLaDA2-Mini-DMax.

\begin{tcolorbox}[contributionbox, title={\faLightbulb~Main Contributions}]
\begin{itemize}[leftmargin=*, itemsep=5pt, label=\textcolor{mbdteal}{\faCheckCircle}]
    \item \textbf{Unified MBD-LM formulation.}
    We formulate \textit{Multi-Block Diffusion Language Models} (MBD-LMs) as a unified DLM framework parameterized by a \textit{running-set} of consecutive blocks.
    This view covers both TF-trained BD-LMs and D2F-trained BD-LMs, while identifying practical MultiBD as the bounded intermediate regime for reliable and efficient inference.

    \item \textbf{MultiTF post-training for MBD-LMs.}
    We propose Multi-block Teacher Forcing (MultiTF), a post-training method that turns BD-LMs into MBD-LMs.
    MultiTF improves train--inference alignment by training BD-LMs on states that resemble practical MultiBD inference.

    \item \textbf{Optimized MultiBD inference engine.}
    We design and implement an optimized MultiBD inference pipeline based on the \textit{Block Buffer} mechanism.
    The pipeline overlaps decoding and KV cache storing, preserves prefix caching, and keeps input shapes static for CUDA Graph capture and replay, translating increased TPF into practical TPS gains.
\end{itemize}
\end{tcolorbox}

%% file: sections/02_preliminary.tex
\section{Preliminaries}

\subsection{Diffusion Language Models}

Diffusion Language Models (DLMs)~\citep{sahoo2024mdlm,nie2025llada,ye2025dream} formulate text generation as iterative denoising.
Let $\mathcal{V}$ denote the vocabulary, $\texttt{[M]}$ denote a special mask token, and $L$ denote the sequence length.
Given a clean sequence $\mathbf{x}_0=(x_0^1,\ldots,x_0^L)\in\mathcal{V}^L$, the forward process gradually masks tokens independently.
For $t\in[0,1]$, the noisy sequence $\mathbf{x}_t\in(\mathcal{V}\cup\{\texttt{[M]}\})^L$ masks each token with probability $t$:
\begin{equation}
q_t(x_t^i \mid x_0^i)=
\begin{cases}
1-t, & x_t^i=x_0^i,\\
t, & x_t^i=\texttt{[M]},\\
0, & \text{otherwise}.
\end{cases}
\label{eq:forward}
\end{equation}
Let $\mathcal{M}(\mathbf{x}_t)=\{i:x_t^i=\texttt{[M]}\}$ denote the masked positions.
A DLM parameterized by $\theta$ predicts clean tokens at masked positions:
\begin{equation}
p_\theta(\mathbf{x}_0 \mid \mathbf{x}_t)
=
\prod_{i=1}^{L}
p_\theta(x_0^i \mid \mathbf{x}_t).
\label{eq:reverse}
\end{equation}
The standard training objective is a weighted masked-token cross-entropy~\citep{nie2025llada}:
\begin{equation}
\begin{aligned}
\mathcal{L}_{\mathrm{DLM}}(\theta)
=
-\mathbb{E}_{t,\mathbf{x}_0,\mathbf{x}_t}
\bigg[
\frac{1}{t}
\sum_{i=1}^{L}
&\mathbf{1}[x_t^i=\texttt{[M]}] \cdot
\log p_\theta(x_0^i\mid \mathbf{x}_t)
\bigg],
\end{aligned}
\label{eq:loss_dllm}
\end{equation}
where $t\sim\mathcal{U}(0,1)$, $\mathbf{x}_t\sim q_t(\cdot\mid\mathbf{x}_0)$, and 
$\mathbf{1}[\cdot]$ denotes the indicator function, ensuring that the loss is computed only on masked tokens.
The inference starts from an all-\texttt{[M]} sequence and iteratively fills high-confidence masked positions.

\begin{figure*}[t]
  \centering
  \includegraphics[width=\textwidth]{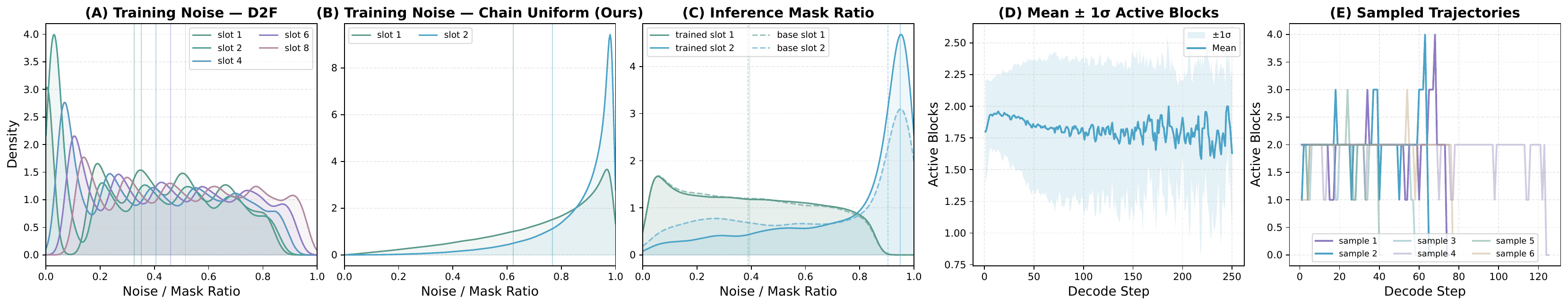}
  \caption{
  Train--inference statistics for MultiBD.
  \textbf{(A)} Slot-wise mask-ratio distributions induced by the D2F-style monotonic scheduler.
  \textbf{(B)} Slot-wise mask-ratio distributions induced by our chain-uniform scheduler.
  \textbf{(C)} Inference-time mask-ratio distributions before and after MultiTF post-training.
  \textbf{(D)} Mean and one-standard-deviation range of the active-block count during MultiBD inference.
  \textbf{(E)} Sampled active-block trajectories during decoding.
  Panels (A--C) compare scheduler-induced training noise patterns with inference-time mask-ratio patterns for train--inference alignment analysis.
  Panels (D--E) report the active part of the MultiBD \textit{running-set} under a buffer size of four; the active-block count can therefore occasionally exceed two.
  }
  \label{fig:train_infer_noise_activity}
\end{figure*}

\subsection{Block Diffusion Language Models}
\label{sec:prelim_bdlm}

Block Diffusion Language Models (BD-LMs)~\citep{arriola2025block,bie2025llada2} partition the sequence into blocks, i.e.,
\begin{equation}
\mathbf{x}_0=[\mathbf{b}_1,\ldots,\mathbf{b}_K],
\qquad
\mathbf{b}_k\in\mathcal{V}^{B},
\end{equation}
where $B$ is the block size and $K=L/B$ is the number of blocks.
BD-LMs model the sequence autoregressively at the block level:
\begin{equation}
\begin{aligned}
p_\theta(\mathbf{x}_0)
&=
\prod_{k=1}^{K}
p_\theta(\mathbf{b}_k\mid \mathbf{x}_0^{(<k)}), \;\mathbf{x}_0^{(<k)}=[\mathbf{b}_1,\ldots,\mathbf{b}_{k-1}].    
\end{aligned}
\label{eq:bdlm_factorization}
\end{equation}
Each conditional term is implemented by a DLM decoding process within the current block.
The block-causal attention pattern is used to allow each block to attend to itself and preceding blocks. 
This enables KV caching during \textit{Single-Block Diffusion} (SingleBD) inference.

\paragraph{Teacher forcing.}
Block Diffusion~\citep{arriola2025block} trains BD-LMs under Teacher Forcing (TF).
For block $\mathbf{b}_k$, only the current block is corrupted by the same masking process,
\begin{equation}
\mathbf{b}_{k,t}\sim q_t(\cdot\mid \mathbf{b}_k),
\end{equation}
and the model predicts masked tokens conditioned on clean prefix blocks:
\begin{equation}
\begin{aligned}
\mathcal{L}_{\mathrm{TF}}(\theta)
=
-\mathbb{E}_{k,t,\mathbf{x}_0,\mathbf{b}_{k,t}}
\bigg[
\frac{1}{t}
\sum_{i=1}^{B}
\mathbf{1}[b_{k,t}^i=\texttt{[M]}]
\cdot
\log p_\theta
\bigl(b_k^i\mid \mathbf{x}_0^{(<k)},\mathbf{b}_{k,t}\bigr)
\bigg].
\end{aligned}
\label{eq:block_diffusion}
\end{equation}
Namely, the model only learns to decode one noisy block conditioned on clean prefix blocks, which is conceptually incompatible with the aforementioned MultiBD inference.

\paragraph{Discrete diffusion forcing.}
Another training paradigm for BD-LMs is Discrete Diffusion Forcing (D2F)~\citep{wang2025d2f}.
D2F introduces visibility among noisy blocks by sampling block-level noise ratios
$\mathbf{t}=(t_1,\ldots,t_K)$ for a block-partitioned suffix.

Let
\[
\mathbf{x}_0^{\mathrm{pre}}
=
(x_0^1,\ldots,x_0^P)\in\mathcal{V}^{P}
\]
denote a clean token-level prefix of length $P$, and let
\[
\mathbf{x}_0^{\mathrm{suf}}
=
[\mathbf{b}_1,\ldots,\mathbf{b}_K],
\qquad
\mathbf{b}_k\in\mathcal{V}^{B},
\]
denote the suffix partitioned into blocks.
D2F constructs noisy suffix blocks
\begin{equation}
\begin{aligned}
\mathbf{x}_{\mathbf{t}}^{\mathrm{suf}}
=
[\mathbf{b}_{1,t_1},\ldots,\mathbf{b}_{K,t_K}],
\quad
\mathbf{b}_{k,t_k}\sim q_{t_k}(\cdot\mid \mathbf{b}_k),
\end{aligned}
\end{equation}
where $0\le t_1<\cdots<t_K\le 1$.
Thus, earlier suffix blocks are less masked, while later suffix blocks are more uncertain.
Conditioned on the clean prefix, D2F trains the student to predict each suffix block from a noisy-prefix view:
\begin{equation}
\begin{aligned}
p_\theta
\bigl(
\mathbf{x}_0^{\mathrm{suf}}
\mid
\mathbf{x}_0^{\mathrm{pre}},
\mathbf{x}_{\mathbf{t}}^{\mathrm{suf}}
\bigr)
=
\prod_{k=1}^{K}
p_\theta
\bigl(
\mathbf{b}_k
\mid
\mathbf{x}_0^{\mathrm{pre}},
\mathbf{b}_{1,t_1},\ldots,\mathbf{b}_{k,t_k}
\bigr).
\end{aligned}
\label{eq:d2f_factorization}
\end{equation}
In practice, D2F is trained with an asymmetric distillation paradigm~\citep{wang2025d2f}.

Despite the goal to perform \textit{Multi-Block Diffusion} (MultiBD), D2F still differs from MultiBD inference in its training states, as detailed in Section~\ref{sec:mbd_lms}.
Beyond the aforementioned mismatch, native D2F also raises a prefix-caching concern.
Its clean prefix $\mathbf{x}_0^{\mathrm{pre}}$ can have arbitrary length $P$ and is processed with full attention rather than block-causal attention.
Therefore, its native formulation is not directly compatible with the prefix caching of BD-LMs.
We analyze this issue in Appendix~\ref{sec:appendix_prefix_cache}, where we compare native D2F with a fully block-causal D2F variant and show that enforcing cache compatibility causes a larger quality degradation, further motivating MultiTF.

%% file: sections/04_methodology.tex
\section{Methodology}
\label{sec:method}

\subsection{Multi-Block Diffusion Language Models}
\label{sec:mbd_lms}

\textit{Multi-Block Diffusion} (MultiBD) generalizes the standard BD-LM factorization in Equation~\ref{eq:bdlm_factorization} by allowing a \textit{running-set} of consecutive blocks to be decoded concurrently.
At decoding step $s$, MultiBD maintains a \textit{running-set}
\[
\mathcal{R}_s=\{a_s,\ldots,c_s\},
\]
where $a_s$ and $c_s$ denote the first and last block indices that have not yet entered the prefix KV cache.
The \textit{running-set} contains the real blocks currently involved in MultiBD decoding, including active noisy blocks and completed preceding blocks waiting to be cached.
Blocks before the \textit{running-set} have already been committed and form the clean cached prefix:
\[
\mathbf{x}_0^{(<a_s)}=[\mathbf{b}_1,\ldots,\mathbf{b}_{a_s-1}].
\]

For each block $k\in\mathcal{R}_s$, let $t_{k,s}\in[0,1]$ denote its current mask ratio at decoding step $s$.
If block $k$ is still active, $\mathbf{b}_{k,t_{k,s}}$ is its current noisy state.
If block $k$ is completed but not yet cached, we set $t_{k,s}=0$, so that $\mathbf{b}_{k,t_{k,s}}=\mathbf{b}_{k,0}=\mathbf{b}_k$.
We refer to each relative block position inside $\mathcal{R}_s$ as a logical slot; for example, the block at index $a_s$ is the first slot and the block at index $a_s+1$ is the second slot.

We define \textit{Multi-Block Diffusion Language Models} (MBD-LMs) as:
\begin{equation}
\begin{aligned}
p_\theta(\mathbf{b}_{\mathcal{R}_s}\mid \mathbf{x}_0^{(<a_s)}, \mathbf{b}_{\mathcal{R}_s,\mathbf{t}_s})
=
\prod_{k=a_s}^{c_s}
p_\theta
\bigl(
\mathbf{b}_k
\mid
\mathbf{x}_0^{(<a_s)},
\mathbf{b}_{a_s,t_{a_s,s}},\ldots,\mathbf{b}_{k,t_{k,s}}
\bigr),
\end{aligned}
\label{eq:multibd_state}
\end{equation}
where
\[
\mathbf{b}_{\mathcal{R}_s}=[\mathbf{b}_{a_s},\ldots,\mathbf{b}_{c_s}],
\qquad
\mathbf{b}_{\mathcal{R}_s,\mathbf{t}_s}
=
[\mathbf{b}_{a_s,t_{a_s,s}},\ldots,\mathbf{b}_{c_s,t_{c_s,s}}].
\]

\begin{wrapfigure}[19]{r}{0.5\columnwidth}
  \centering
  \begin{adjustbox}{width=\linewidth}
  \includegraphics{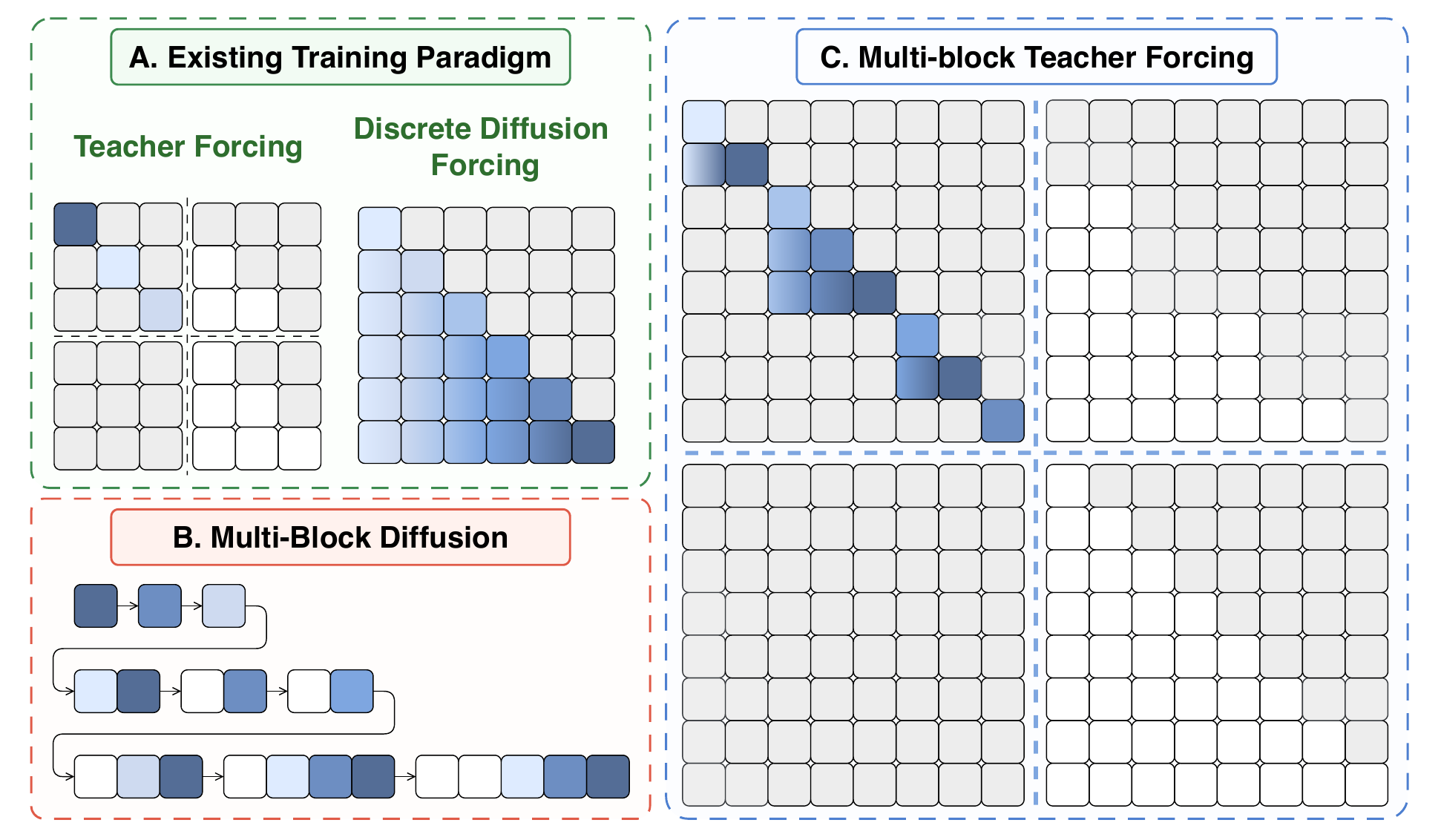}
  \end{adjustbox}
  \caption{
    Train--inference alignment across paradigms.
    \textbf{(A)} TF and D2F provide existing BD-LM training states, but neither matches practical MultiBD.
    \textbf{(B)} MultiBD maintains a bounded \textit{running-set} for concurrent block refinement.
    \textbf{(C)} MultiTF builds inference-like \textit{noise-groups} with heterogeneous slot-wise noise patterns.
    }
  \label{fig:train_inference_match}
\end{wrapfigure}

This formulation asks the model to recover the current \textit{running-set} from the clean cached prefix and the visible block states inside $\mathcal{R}_s$.
The \textit{running-set} size is defined as $|\mathcal{R}_s|$.

The \textit{running-set} view gives a unified way to describe existing BD-LM regimes.
As illustrated in Figure~\ref{fig:train_inference_match}, TF-trained BD-LMs correspond to the SingleBD extreme, where the model only observes one noisy block conditioned on a clean cached prefix.
D2F-trained BD-LMs introduce visibility among multiple noisy suffix blocks, but their training states still differ from practical MultiBD inference in \textit{running-set} size and slot-wise noise patterns.
Under the MBD-LM formulation, these regimes can be viewed as limiting cases, while practical MultiBD is the bounded intermediate regime that decodes a small \textit{running-set} concurrently.

Conceptually, MultiBD reduces to SingleBD when $|\mathcal{R}_s|=1$: the model decodes only one block conditioned on the clean cached prefix.
At the other extreme, if the \textit{running-set} is expanded to cover all suffix blocks and a monotonic D2F-style \textit{noise-scheduler} is used, the resulting training state resembles the fully block-causal D2F variant discussed in Appendix~\ref{sec:appendix_prefix_cache}.
This connection is only at the level of training-state construction: D2F remains a training paradigm, while MultiBD is the inference regime targeted by MBD-LMs.
In practice, useful MultiBD operates between these two extremes:
$|\mathcal{R}_s|$ should be larger than $1$ to expose inter-block parallelism, but remain bounded to keep each forward pass efficient and executable.
This bounded \textit{running-set} view is consistent with the empirical MultiBD traces analyzed in Section~\ref{sec:train_inference_alignment}, and is reflected in both the training-side and inference-side designs proposed below.

\subsection{Multi-block Teacher Forcing}
\label{sec:training}

Multi-block Teacher Forcing (MultiTF) post-trains BD-LMs into MBD-LMs by constructing inference-like training states, with particular emphasis on matching the bounded \textit{running-set} structure and the slot-wise noise patterns of MultiBD inference.
MultiTF can be viewed as an extension of TF from one noisy block to a bounded group of consecutive noisy blocks.
We call such a group a \textit{noise-group}.
Following the bounded \textit{running-set} view in Section~\ref{sec:mbd_lms}, MultiTF uses $G_{\max}$ as the training-side upper bound on \textit{noise-group} size.
Throughout the paper, $G_{\max}$ denotes the maximum \textit{noise-group} size, $\Lambda$ denotes the set of sampled \textit{group-layouts}, $\lambda\in\Lambda$ denotes one layout, and $H_m$ denotes one \textit{noise-group}.
Each \textit{noise-group} $H_m$ is constructed as a bounded training analogue of a possible MultiBD \textit{running-set}.
Notably, later \textit{noise-groups} are conditioned on \emph{clean} earlier \textit{noise-groups} during training.

\begin{figure*}[t]
  \centering
  \includegraphics[width=\textwidth]{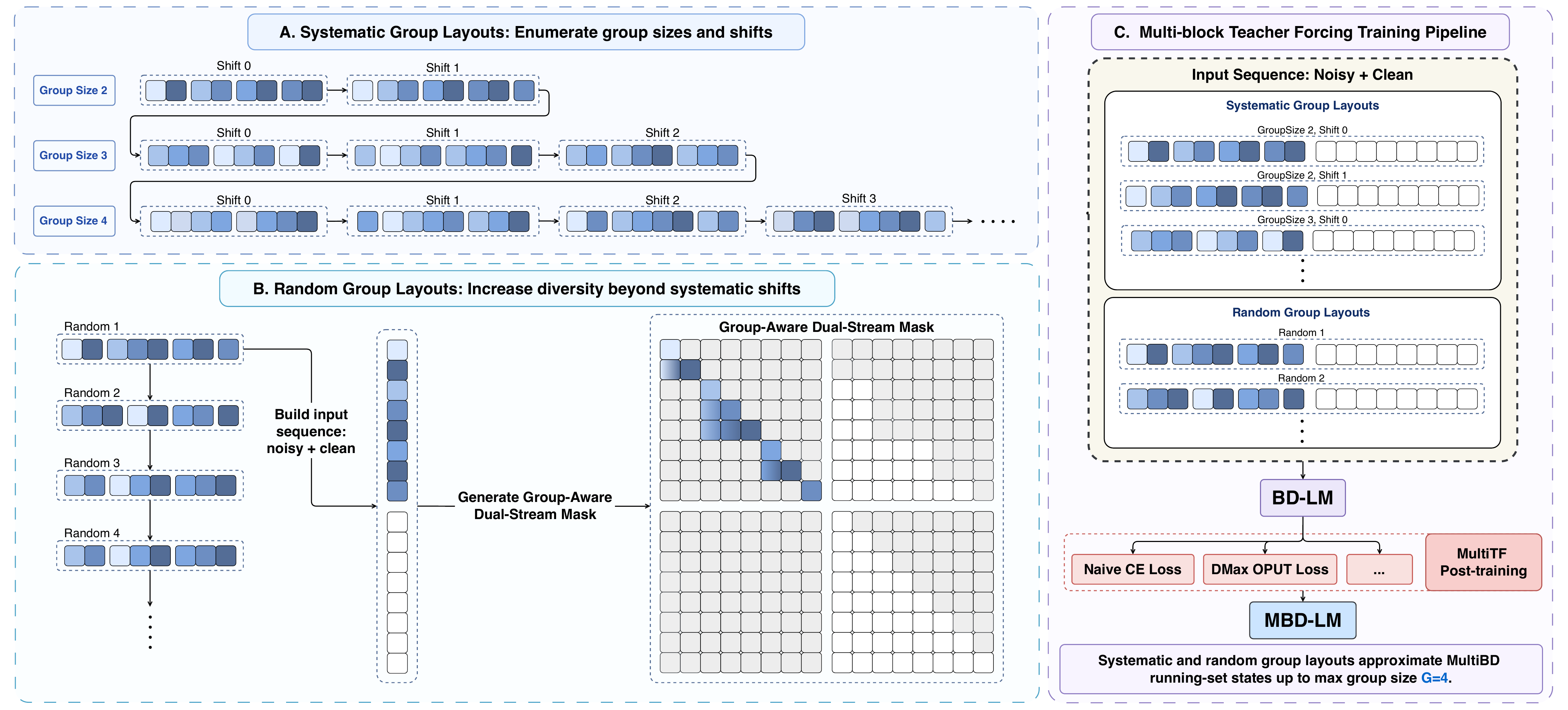}
    \caption{
    Overview of MultiTF.
    \textbf{(A)} Systematic \textit{group-layouts} enumerate group sizes and shifts so that blocks appear at different group-relative positions.
    \textbf{(B)} Random \textit{group-layouts} increase layout diversity; each layout is converted into a noisy--clean input sequence with the \textit{Group-Aware Dual-Stream Mask}.
    \textbf{(C)} The resulting input sequences are used to post-train BD-LMs into MBD-LMs with masked CE and optional model-specific objectives.
    }
  \label{fig:multibd_training_pipeline}
\end{figure*}

\begin{algorithm}[t]
\footnotesize
\caption{Multi-block Teacher Forcing}
\label{alg:multitf_training}
\begin{algorithmic}[1]
\Require Clean sequence $\mathbf{x}_0$; block size $B$; maximum \textit{noise-group} size $G_{\max}$; noise bounds $t_{\mathrm{low}}, t_{\mathrm{high}}$; margin ratio $\rho$; number of random layouts $N_{\mathrm{rand}}$; mask token $\texttt{[M]}$.

\Statex \textcolor{commentcolor}{// \textbf{Construct \textit{noise-group} layouts}}
\State Partition $\mathbf{x}_0$ into $K$ blocks $[\mathbf{b}_1,\ldots,\mathbf{b}_K]$.
\State Generate systematic layouts by enumerating \textit{noise-group} sizes $g\in\{2,\ldots,G_{\max}\}$ and all $g$ group shifts.
\State Generate $N_{\mathrm{rand}}$ random layouts by sampling \textit{noise-group} sizes from $\{2,\ldots,G_{\max}\}$ until all blocks are covered.
\State Let $\Lambda$ be the union of systematic and random layouts.

\Statex \textcolor{commentcolor}{// \textbf{Apply MultiTF corruption and training}}
\State Set $t_{\mathrm{eff}}\gets t_{\mathrm{high}}-\rho(t_{\mathrm{high}}-t_{\mathrm{low}})$, where $\rho$ is the noise-transition margin ratio.
\State Initialize accumulated loss $\mathcal{J}\gets 0$.
\For{each layout $\lambda\in\Lambda$}
    \State Initialize noisy sequence $\mathbf{x}^{\lambda}_{\mathbf{t}}\gets\mathbf{x}_0$.
    \For{each \textit{noise-group} $H_m=(j_1,\ldots,j_{n_m})\in\lambda$}
        \Statex \hspace{\algorithmicindent}\hspace{\algorithmicindent}\textcolor{commentcolor}{// \textbf{Chain-uniform block-level \textit{noise-scheduler}}}
        \State Sample group floor $\ell\sim\mathcal{U}(t_{\mathrm{low}},t_{\mathrm{eff}})$.
        \For{$i\gets 1$ \textbf{to} $n_m$}
            \State Sample $t_{j_i}\sim\mathcal{U}(\ell,t_{\mathrm{eff}})$ and set $\ell\gets t_{j_i}$.
            \State Mask $\lfloor B\cdot t_{j_i}\rfloor$ random positions in $\mathbf{b}_{j_i}$ as $\texttt{[M]}$.
        \EndFor
    \EndFor

    \Statex \hspace{\algorithmicindent}\textcolor{commentcolor}{// \textbf{Build input sequence and attention mask}}
    \State Construct $\mathbf{X}_{\lambda}=[\mathbf{x}^{\lambda}_{\mathbf{t}};\mathbf{x}_0]$.
    \State Construct the \textit{Group-Aware Dual-Stream Mask} $\mathbf{A}_{\lambda}$.
    \State Run the model on $(\mathbf{X}_{\lambda},\mathbf{A}_{\lambda})$.

    \Statex \hspace{\algorithmicindent}\textcolor{commentcolor}{// \textbf{Compute masked CE}}
    \State Let $\mathcal{M}_{\lambda}=\{i:\mathbf{x}^{\lambda}_{\mathbf{t}}[i]=\texttt{[M]}\}$.
    \State Compute layout-level masked CE estimate $\mathcal{J}_{\lambda}$ over $\mathcal{M}_{\lambda}$.
    \State $\mathcal{J}\gets\mathcal{J}+\mathcal{J}_{\lambda}$.
\EndFor
\State \Return $\mathcal{J}/|\Lambda|$.
\end{algorithmic}
\end{algorithm}

\begin{figure*}[t]
  \centering
  \includegraphics[width=0.95\textwidth]{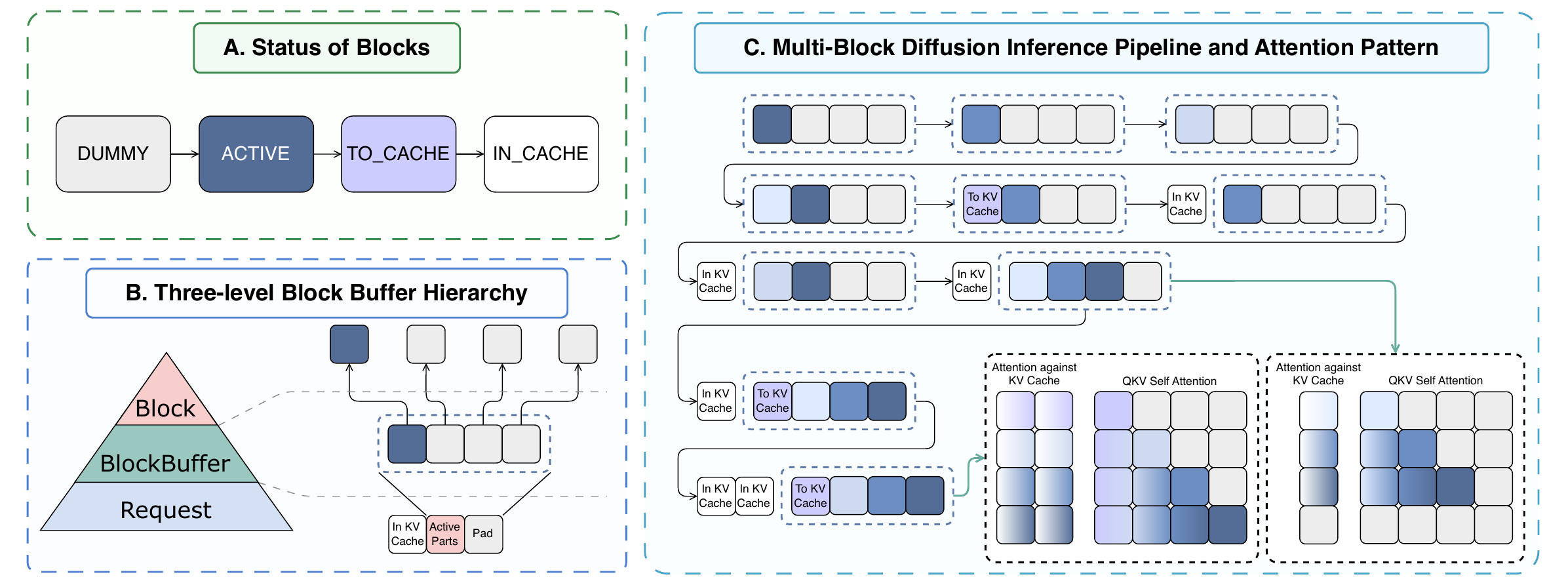}
  \caption{
  Inference and system support in MultiBD.
  \textbf{(1)} Blocks follow a four-state transition:
  \textsc{dummy} $\rightarrow$ \textsc{active} $\rightarrow$ \textsc{to-cache} $\rightarrow$ \textsc{in-cache}.
  \textbf{(2)} MultiBD organizes decoding with a block--buffer--request hierarchy, where each request maintains \textit{Block Buffers} and each buffer contains multiple block slots for parallel refinement.
  \textbf{(3)} During MultiBD inference, noisy blocks are refined jointly under block-causal self-attention, while committed prefix blocks are served from the KV cache; completed blocks enter the cache and the \textit{Block Buffer} slides forward.
  }
  \label{fig:multibd_inference_pipeline}
\end{figure*}

Here $\mathcal{J}$ is only the finite-layout estimator accumulated inside Algorithm~\ref{alg:multitf_training}; the population-level training objective is $\mathcal{L}_{\mathrm{MultiTF}}$ in Equation~\ref{eq:multitf_loss}.

\paragraph{Group-layout construction.}
Given a clean block sequence $[\mathbf{b}_1,\ldots,\mathbf{b}_K]$, MultiTF constructs a set of \textit{group-layouts}
$\Lambda$, where each \textit{group-layout} $\lambda=(H_1,\ldots,H_{|\lambda|})$ partitions the sequence into consecutive \textit{noise-groups}.
Each \textit{noise-group} $H_m=\{a_m,\ldots,c_m\}$ has the same consecutive-block form as a possible MultiBD \textit{running-set}
$\mathcal{R}_s=\{a_s,\ldots,c_s\}$.
We use both systematic and random \textit{group-layouts} to cover different bounded \textit{running-set} sizes and group-relative positions, as shown in Figure~\ref{fig:multibd_training_pipeline}.

\begin{itemize}
    \item \textbf{Systematic layouts.}
    We specify a maximum \textit{noise-group} size $G_{\max}$.
    For each \textit{noise-group} size $g\in\{2,\ldots,G_{\max}\}$ and each shift $h\in\{0,\ldots,g-1\}$, we define a shifted layout
    $\lambda_{g,h}$ by placing group boundaries every $g$ blocks with offset $h$:
    \[
    H_{g,h,q}
    =
    \{\,1+h+qg,\ldots,h+(q+1)g\,\}\cap\{1,\ldots,K\},
    \]
    where $q$ indexes groups within the shifted layout, and boundary groups are clipped to the valid block range.
    The systematic layout set is
    \[
    \Lambda_{\mathrm{sys}}
    =
    \{\lambda_{g,h}: g\in\{2,\ldots,G_{\max}\},\; h\in\{0,\ldots,g-1\}\}.
    \]
    This construction ensures that, ignoring boundary effects, every consecutive \textit{running-set}
    $\{a,\ldots,a+g-1\}$ of length $g$ appears as one \textit{noise-group} in exactly one shifted layout, with shift
    $h=(a-1)\bmod g$.
    Equivalently, for each fixed $g$, every block appears once at every group-relative position across the $g$ shifts.

    \item \textbf{Random layouts.}
    Systematic layouts provide structured coverage but are regular by construction.
    To increase layout diversity, we further sample random layouts by drawing \textit{noise-group} sizes
    $g_m\in\{2,\ldots,G_{\max}\}$ and forming consecutive groups
    \[
    H_m=\{a_m,\ldots,\min(a_m+g_m-1,K)\},
    \qquad
    a_{m+1}=\min(a_m+g_m,K+1),
    \]
    until the full sequence is covered.
    These random layouts add non-regular \textit{noise-group}-size combinations and boundary patterns without replacing the coverage guarantee of systematic layouts.
\end{itemize}

The final layout set is
\[
\Lambda=\Lambda_{\mathrm{sys}}\cup\Lambda_{\mathrm{rand}}.
\]
We provide a theoretical coverage view in Appendix~\ref{sec:appendix_theory}, showing how systematic shifts cover bounded \textit{running-sets} while random layouts add distributional diversity.

\paragraph{Chain-uniform noise-scheduling.}
After sampling a \textit{group-layout}, MultiTF assigns mask ratios within each \textit{noise-group}.
Unlike D2F's monotonic block-level schedule over a long noisy sequence, MultiTF uses a randomized chain-uniform \textit{noise-scheduler} inside each bounded \textit{noise-group}.
Specifically, for each \textit{noise-group}, we sample a group-level floor and then sample each block's mask ratio with the previous block's ratio as the lower bound, as shown in Algorithm~\ref{alg:multitf_training}.
This produces monotonic but randomized group-internal noise levels, encouraging larger slot-wise noise gaps that better match MultiBD inference.

\paragraph{Group-Aware Dual-Stream Mask.}
For each layout $\lambda$, the sampled block-level mask ratios corrupt the clean sequence into a noisy sequence $\mathbf{x}^{\lambda}_{\mathbf{t}}$.
Following the TF-style construction~\citep{arriola2025block}, MultiTF builds the input sequence by concatenating the noisy sequence with the clean sequence:
\begin{equation}
    \mathbf{X}_{\lambda}
    =[\mathbf{x}^{\lambda}_{\mathbf{t}};\mathbf{x}_0].
\end{equation}
The noisy part represents the MultiBD-like decoding state, while the clean part provides clean-prefix context.

We construct a \textit{Group-Aware Dual-Stream Mask} over $\mathbf{X}_{\lambda}$:
\begin{equation}
\mathbf{A}_{\lambda}
=
\begin{bmatrix}
\mathbf{M}_{\mathrm{GD}} & \mathbf{M}_{\mathrm{GOC}} \\
0 & \mathbf{M}_{\mathrm{BC}}
\end{bmatrix},
\label{eq:multitf_attention_mask}
\end{equation}
where $\mathbf{M}_{\mathrm{GD}}$ enables group-internal noisy-block visibility, $\mathbf{M}_{\mathrm{GOC}}$ lets each \textit{noise-group} condition on its clean prefix, and $\mathbf{M}_{\mathrm{BC}}$ preserves standard block-causal visibility on the clean part.
The zero lower-left block prevents clean tokens from attending to noisy tokens; detailed mask definitions are provided in Appendix~\ref{sec:appendix_training}.

\paragraph{Training objective.}
MultiTF optimizes masked-token cross-entropy on the noisy part of the input sequence:
\begin{equation}
\begin{aligned}
\mathcal{L}_{\mathrm{MultiTF}}
=-\mathbb{E}_{\lambda,\mathbf{t},\mathbf{x}_0}
\bigg[
\frac{1}{|\mathcal{M}_{\lambda}|}
\sum_{i\in\mathcal{M}_{\lambda}}
\log p_\theta
\bigl(
x_0^i
\mid
\mathbf{X}_{\lambda},
\mathbf{A}_{\lambda}
\bigr)
\bigg],
\end{aligned}
\label{eq:multitf_loss}
\end{equation}
where
\begin{equation}
\mathcal{M}_{\lambda}
=
\{i:\mathbf{x}^{\lambda}_{\mathbf{t}}[i]=\texttt{[M]}\}
\end{equation}
denotes masked positions on the noisy part.
All systematic and random layouts are batched as independent input sequences, as illustrated in Figure~\ref{fig:multibd_training_pipeline}.
For models with additional objectives, such as DMax, we apply the corresponding model-specific loss on top of the same MultiTF inputs.

The concrete MultiTF objective and model-specific training variants are detailed in Appendix~\ref{sec:appendix_objectives}.

\subsection{Optimized Multi-Block Diffusion}
\label{sec:decoding}

After MultiTF post-training, an MBD-LM performs MultiBD inference over the \textit{running-set} $\mathcal{R}_s$ in Equation~\ref{eq:multibd_state}.
The inference objective is to expose inter-block parallelism without losing the serving advantages of BD-LMs.
Concretely, practical MultiBD should satisfy the following inference requirements:
\begin{tcolorbox}[requirementbox, title=\textit{Inference Requirements for Practical MultiBD}]
\begin{itemize}[leftmargin=*, itemsep=3pt]
    \item \textbf{Inter-block parallelism:} multiple noisy blocks are decoded in parallel.
    \item \textbf{Decode-store overlap:} decoding of later active blocks overlaps with KV cache storing of completed preceding blocks.
    \item \textbf{Prefix-cache preservation:} committed prefix blocks should produce stable KV cache that remains reusable by the standard BD-LM prefix cache.
    \item \textbf{Static-shape execution:} the physical input shape remains fixed for CUDA Graph capture and replay and efficient execution.
\end{itemize}
\end{tcolorbox}

\paragraph{Naive MultiBD and dynamic execution.}
A naive block-causal MultiBD implementation naturally supports inter-block parallelism and decode-store overlap.
As illustrated in Figure~\ref{fig:single_vs_multiblock_pipeline} and detailed in Algorithm~\ref{alg:naive_multibd}, it directly materializes the \textit{running-set} $\mathcal{R}_s$ as the input to each forward pass:
future noisy blocks are appended to $\mathcal{R}_s$ when the latest active block makes sufficient progress, and completed preceding blocks are removed after being cached.
Thus, later blocks can already be decoded while earlier completed blocks are being stored, avoiding the storing bubbles of SingleBD.
This dynamic procedure only needs three logical block states,
\[
\textsc{active} \rightarrow \textsc{to-cache} \rightarrow \textsc{in-cache},
\]
because every block in the \textit{running-set} corresponds to a real block being decoded or committed.
However, since each forward pass is built directly from $\mathcal{R}_s$, the number of processed tokens changes over time and across requests, making CUDA Graph capture and replay difficult.

\paragraph{Static-shape execution with \textit{Block Buffer}.}
To satisfy all four requirements simultaneously, we decouple the logical \textit{running-set} from the physical input by using a \textit{Block Buffer} mechanism, as detailed in Algorithm~\ref{alg:optimized_multibd}.
As shown in Figure~\ref{fig:multibd_inference_pipeline}(B), our inference engine organizes MultiBD decoding with a three-level hierarchy: a request manages one or more \textit{Block Buffers}, each \textit{Block Buffer} contains a fixed number of block slots, and each slot stores one block state.
The request level handles generation progress and cache ownership, the \textit{Block Buffer} level provides a static physical input for CUDA Graph replay, and the block level tracks whether each slot is \textsc{dummy}, \textsc{active}, \textsc{to-cache}, or \textsc{in-cache}.

Let $\mathcal{W}_s$ denote the physical \textit{Block Buffer} at decoding step $s$.
It contains a fixed number of block slots:
\[
|\mathcal{W}_s|=N_{\mathrm{buf}},
\]
where $N_{\mathrm{buf}}$ is the buffer size.
The real resident blocks inside $\mathcal{W}_s$ form the \textit{running-set} $\mathcal{R}_s$, while the remaining slots are dummy slots.
Thus, the buffer can be written as
\[
\mathcal{W}_s=\mathcal{R}_s\Vert\mathcal{D}_s,
\qquad
|\mathcal{W}_s|=|\mathcal{R}_s|+|\mathcal{D}_s|=N_{\mathrm{buf}},
\qquad
|\mathcal{R}_s|\le N_{\mathrm{buf}},
\]
where $\mathcal{D}_s$ denotes the trailing dummy segment.
Thus, $N_{\mathrm{buf}}$ is the inference-side realization of the bounded \textit{running-set} assumption introduced in Section~\ref{sec:mbd_lms}.
In practice, $N_{\mathrm{buf}}$ is chosen within the \textit{running-set} sizes covered by MultiTF through $G_{\max}$.

A future block enters decoding by activating an existing dummy slot rather than extending the physical input sequence.
When the front block of $\mathcal{R}_s$ is completed, it is marked as \textsc{to-cache}; once committed to the KV cache, it leaves $\mathcal{R}_s$ and becomes part of the cached prefix.
The \textit{Block Buffer} then slides forward by appending a new dummy slot at the tail.
Thus, MultiBD can advance its \textit{running-set} while keeping the physical buffer shape fixed, thereby enabling static-shape execution for CUDA Graph capture and replay.

As shown in Figure~\ref{fig:multibd_inference_pipeline}(A), each physical slot follows the state transition
\[
\textsc{dummy} \rightarrow \textsc{active} \rightarrow \textsc{to-cache} \rightarrow \textsc{in-cache}.
\]
The key difference from the naive three-state dynamic procedure is the additional \textsc{dummy} state, which reserves inactive capacity inside the \textit{Block Buffer}.
This allows future blocks to enter by activating existing slots instead of extending the physical input, while completed front blocks are committed into the KV cache.

\paragraph{Prefix-cache preservation.}
The \textit{Block Buffer} mechanism also preserves the cache semantics of block-causal BD-LMs.
Committed front blocks become immutable clean prefix blocks and are represented only through cached KV states, while active blocks remain inside the \textit{Block Buffer} for iterative refinement.
This separation is important because native D2F uses prefix-full attention and is not directly compatible with the standard BD-LM prefix-cache interface, as discussed in Section~\ref{sec:prelim_bdlm}.
Appendix~\ref{sec:appendix_prefix_cache} further shows that simply converting D2F into a fully block-causal variant improves cache compatibility but causes a larger quality degradation.
In contrast, MultiTF trains MBD-LMs with block-causal clean-prefix conditioning, and the \textit{Block Buffer} inference pipeline preserves this prefix-cache interface during MultiBD decoding.

This design preserves inter-block parallelism, overlaps decoding with KV cache storing, maintains prefix-cache reuse, and supports static-shape execution for CUDA Graph replay.
As a result, the increased TPF of MBD-LMs can be converted into practical wall-clock speedup.
Additional implementation details, including the naive dynamic MultiBD, the optimized MultiBD, block-state transitions, threshold rules, and prefix-cache analysis, are provided in Appendix~\ref{sec:appendix_system}.
The realized speedup is validated by the TPS results in Table~\ref{tab:tps_comparison}.

%% file: sections/05_experiments.tex
\section{Experiments}

\subsection{Experimental Setup}

\noindent{\textbf{Models and training.}}
We evaluate MultiTF on representative BD-LMs from the LLaDA2.x~\citep{bie2025llada2,bie2026llada21} and SDAR~\citep{cheng2025sdar} families, including variants enhanced with DMax~\citep{chen2026dmax}.
For each base model, MultiTF post-training constructs multiple \textit{group-layouts} per sample, including systematic shifted layouts and random layouts, to approximate the MultiBD \textit{running-set} states described in Section~\ref{sec:mbd_lms}.
The resulting models are denoted as MBD-* models, e.g., MBD-LLaDA2-Mini and MBD-SDAR-8B-Chat.
We also evaluate training-free MultiBD, which directly applies MultiBD inference to the original BD-LMs without post-training.

\begin{table*}[t]
\centering
\caption{\textbf{Evaluation results across math and code benchmarks.} SingleBD (Native) denotes the native single-block diffusion inference of each BD-LM; MultiBD (training-free) applies multi-block decoding without retraining; MBD-* denotes the corresponding MultiTF-post-trained MBD-LM. AUP (Accuracy Under Parallelism) combines accuracy and TPF, reported in the \textbf{Average} column as an aggregate across four benchmarks. MBD-LMs consistently improve TPF over SingleBD. In most settings, MultiTF recovers or improves the quality lost by training-free MultiBD, leading to a better accuracy--parallelism trade-off.}
\label{tab:main_results}
\resizebox{\textwidth}{!}{%
\begin{tabular}{lccccccccccc}
\toprule
& \multicolumn{2}{c}{\textbf{GSM8K}} & \multicolumn{2}{c}{\textbf{MATH500}} & \multicolumn{2}{c}{\textbf{MBPP+}} & \multicolumn{2}{c}{\textbf{HumanEval+}} & \multicolumn{3}{c}{\textbf{Average}} \\
\cmidrule(lr){2-3} \cmidrule(lr){4-5} \cmidrule(lr){6-7} \cmidrule(lr){8-9} \cmidrule(lr){10-12}
\textbf{Model} 
& \textbf{Acc $\uparrow$} & \textbf{TPF $\uparrow$} 
& \textbf{Acc $\uparrow$} & \textbf{TPF $\uparrow$} 
& \textbf{Acc $\uparrow$} & \textbf{TPF $\uparrow$} 
& \textbf{Acc $\uparrow$} & \textbf{TPF $\uparrow$} 
& \textbf{Acc $\uparrow$} & \textbf{TPF $\uparrow$} & \textbf{AUP $\uparrow$} \\
\midrule

\multicolumn{12}{l}{\textbf{\textit{LLaDA2-Mini-DMax}} \textit{(bufsz=2, blksz=32)}} \\
\hspace{1em}SingleBD (Native)
& \textbf{91.89} & 5.70
& \textbf{76.80} & 6.13
& \textbf{72.22} & 6.14
& \textbf{77.44} & 7.44
& \textbf{79.59} & 6.35 & 459.54 \\
\rowcolor{trainingfreegreen}
\hspace{1em}MultiBD (training-free)
& 89.84 & \underline{8.76}
& 73.80 & \underline{9.08}
& \textbf{72.22} & \textbf{8.44}
& \underline{76.83} & \textbf{10.96}
& 78.17 & \underline{9.31} & \underline{651.98} \\
\rowcolor{multibdblue}
\hspace{1em}\textbf{MBD-LLaDA2-Mini-DMax}
& \underline{91.74} & \textbf{8.95}
& \underline{75.00} & \textbf{9.31}
& \underline{70.11} & \underline{8.34}
& \textbf{77.44} & \underline{10.78}
& \underline{78.57} & \textbf{9.34} & \textbf{661.28} \\
\midrule

\multicolumn{12}{l}{\textbf{\textit{LLaDA2-Mini}} \textit{(bufsz=2, blksz=32)}} \\
\hspace{1em}SingleBD (Native)
& 91.89 & 2.27
& \underline{74.20} & 2.83
& \textbf{75.66} & 3.25
& \underline{78.05} & 5.53
& \underline{79.95} & 3.47 & 247.41 \\
\rowcolor{trainingfreegreen}
\hspace{1em}MultiBD (training-free)
& \textbf{92.65} & \underline{2.76}
& 73.60 & \underline{3.53}
& \underline{72.49} & \underline{3.97}
& 75.61 & \underline{7.37}
& 78.59 & \underline{4.41} & \underline{301.81} \\
\rowcolor{multibdblue}
\hspace{1em}\textbf{MBD-LLaDA2-Mini}
& \underline{91.96} & \textbf{5.55}
& \textbf{79.20} & \textbf{6.02}
& \underline{72.49} & \textbf{5.35}
& \textbf{80.49} & \textbf{7.85}
& \textbf{81.03} & \textbf{6.19} & \textbf{449.18} \\
\midrule

\multicolumn{12}{l}{\textbf{\textit{SDAR-8B-Chat-b32}} \textit{(bufsz=4, blksz=32)}} \\
\hspace{1em}SingleBD (Native)
& \textbf{90.07} & 2.52
& \underline{65.60} & 3.81
& \underline{52.65} & 1.83
& \textbf{67.68} & 2.00
& \underline{69.00} & 2.54 & 141.64 \\
\rowcolor{trainingfreegreen}
\hspace{1em}MultiBD (training-free)
& 89.01 & \underline{2.78}
& 60.60 & \underline{5.06}
& 52.12 & \underline{1.97}
& \underline{65.85} & \underline{2.24}
& 66.89 & \underline{3.01} & \underline{156.35} \\
\rowcolor{multibdblue}
\hspace{1em}\textbf{MBD-SDAR-8B-Chat-b32}
& \underline{89.16} & \textbf{3.08}
& \textbf{68.00} & \textbf{5.08}
& \textbf{58.99} & \textbf{4.87}
& 62.80 & \textbf{4.82}
& \textbf{69.74} & \textbf{4.46} & \textbf{210.42} \\
\midrule

\multicolumn{12}{l}{\textbf{\textit{SDAR-8B-Chat-b4}} \textit{(bufsz=4, blksz=4)}} \\
\hspace{1em}SingleBD (Native)
& \underline{91.05} & 1.33
& \textbf{72.80} & 1.46
& \underline{64.80} & 1.13
& \underline{73.70} & 1.07
& \textbf{75.59} & 1.25 & 85.46 \\
\rowcolor{trainingfreegreen}
\hspace{1em}MultiBD (training-free)
& 90.45 & \textbf{2.39}
& 70.60 & \textbf{2.68}
& \textbf{65.80} & \underline{1.55}
& \textbf{74.39} & \underline{1.47}
& \underline{75.31} & \underline{2.00} & \underline{129.59} \\
\rowcolor{multibdblue}
\hspace{1em}\textbf{MBD-SDAR-8B-Chat-b4}
& \textbf{91.81} & \underline{2.28}
& \underline{72.40} & \underline{2.52}
& 64.29 & \textbf{2.62}
& 72.56 & \textbf{2.24}
& 75.27 & \textbf{2.42} & \textbf{148.65} \\

\bottomrule
\end{tabular}%
}
\end{table*}

\begin{table*}[t]
\centering

\begin{subtable}[t]{0.55\textwidth}
\centering
\caption{Training-free MultiBD transfers to additional model variants. SingleBD (Native) denotes each model's native single-block diffusion inference.}
\label{tab:transfer_math}
\resizebox{\linewidth}{!}{%
\begin{tabular}{lccccccc}
\toprule
& \multicolumn{2}{c}{\textbf{GSM8K}} & \multicolumn{2}{c}{\textbf{MATH500}} & \multicolumn{3}{c}{\textbf{Average}} \\
\cmidrule(lr){2-3} \cmidrule(lr){4-5} \cmidrule(lr){6-8}
& \textbf{Acc $\uparrow$} & \textbf{TPF $\uparrow$} 
& \textbf{Acc $\uparrow$} & \textbf{TPF $\uparrow$} 
& \textbf{Acc $\uparrow$} & \textbf{TPF $\uparrow$} & \textbf{AUP $\uparrow$} \\
\midrule
\multicolumn{8}{l}{\textit{\textbf{LLaDA2-Mini-CAP} (bufsz=2, blksz=32)}} \\
\hspace{1em}SingleBD (Native)
& \textbf{91.74} & 3.08
& \textbf{77.80} & 3.71
& \textbf{84.77} & 3.40 & 247.30 \\
\rowcolor{trainingfreegreen}
\hspace{1em}MultiBD (training-free)
& 91.21 & \textbf{4.00}
& 77.20 & \textbf{4.94}
& 84.21 & \textbf{4.47} & \textbf{319.17} \\
\midrule
\multicolumn{8}{l}{\textit{\textbf{LLaDA2.1-Mini} (bufsz=2, blksz=32)}} \\
\hspace{1em}SingleBD (Native)
& \textbf{93.03} & 4.12
& \textbf{81.40} & 4.87
& \textbf{87.22} & 4.50 & 390.64 \\
\rowcolor{trainingfreegreen}
\hspace{1em}MultiBD (training-free)
& 92.27 & \textbf{5.80}
& 81.00 & \textbf{7.20}
& 86.63 & \textbf{6.50} & \textbf{558.52} \\
\bottomrule
\end{tabular}%
}
\end{subtable}
\hfill
\begin{subtable}[t]{0.4\textwidth}
\centering
\caption{
Ablation of MultiTF training components averaged over HumanEval+ and GSM8K with LLaDA2-Mini-DMax.
}
\label{tab:ablation}
\setlength{\tabcolsep}{4pt}
\renewcommand{\arraystretch}{0.92}
\resizebox{\linewidth}{!}{%
\begin{tabular}{@{}lccc@{}}
\toprule
\textbf{Configuration} & \textbf{Acc $\uparrow$} & \textbf{TPF $\uparrow$} & \textbf{AUP $\uparrow$} \\
\midrule
SingleBD (Native) & \textbf{84.67} & 6.57 & 536.89 \\
\midrule
\multicolumn{4}{l}{\textbf{\textit{noise-group layouts construction}}} \\
+ systematic layouts & 83.22 & 9.71 & \underline{774.03} \\
+ random layouts & 82.72 & 9.42 & 747.46 \\
\rowcolor{multibdblue}
\textbf{systematic + random layouts (ours)} & \underline{84.59} & \textbf{9.87} & \textbf{805.34} \\
\midrule
\multicolumn{4}{l}{\textbf{\textit{block-level noise-scheduler}}} \\
D2F-style monotonic scheduler & 79.34 & 8.76 & 657.74 \\
random scheduler & 83.14 & 9.70 & 771.74 \\
sorted-uniform scheduler & 81.28 & \underline{9.73} & 748.73 \\
\rowcolor{multibdblue}
\textbf{chain-uniform scheduler (ours)} & \underline{84.59} & \textbf{9.87} & \textbf{805.34} \\
\bottomrule
\end{tabular}%
}
\end{subtable}

\caption{
Transfer and ablation results.
(a) Training-free MultiBD transfers to additional model variants on math benchmarks.
(b) MultiTF component ablations averaged over HumanEval+ and GSM8K.
All reported metrics are higher-is-better.
}
\label{tab:transfer_and_ablation}
\end{table*}

\noindent{\textbf{Benchmarks and metrics.}}
We evaluate mathematical reasoning on GSM8K~\citep{gsm8k} and MATH500~\citep{math}, and code generation on MBPP+ and HumanEval+~\citep{evalplus}.
We report Accuracy, Tokens Per Forward pass (TPF), and Accuracy Under Parallelism (AUP).
Accuracy is exact match for math and pass@1 for code.
TPF measures decoding parallelism, while AUP summarizes the accuracy--parallelism trade-off following d3LLM~\citep{qian2026d3llm}. Given a set of decoding configurations $\mathcal{C}$, we sort them by TPF and compute AUP as the trapezoidal area under the accuracy--TPF curve:
\begin{equation}
\mathrm{AUP}=
\sum_{i=1}^{|\mathcal{C}|-1}
\frac{A_{c_i}+A_{c_{i+1}}}{2}
\left(P_{c_{i+1}}-P_{c_i}\right),
\end{equation}
where $A_{c_i}$ and $P_{c_i}$ denote the accuracy and TPF of configuration $c_i$, respectively. For multi-benchmark evaluation, we report the average AUP across benchmarks.

\noindent{\textbf{Experimental details.}}
Detailed training hyperparameters, inference hyperparameters, hardware settings, and training costs are provided in Appendix~\ref{sec:appendix_exp_details}.

\subsection{Main Results}

We first evaluate whether MBD-LMs can improve decoding parallelism without sacrificing generation quality.
The analysis focuses on four questions:
(i) whether MultiTF-post-trained MBD-LMs improve the TPF--accuracy trade-off over native SingleBD;
(ii) whether MultiTF is complementary to T2T-enhanced decoding methods such as DMax;
(iii) whether train--inference alignment is necessary beyond training-free MultiBD; and
(iv) whether the gains generalize across different BD-LM backbones.

\paragraph{Baselines and configurations.}
Table~\ref{tab:main_results} reports results across four benchmarks.
For each base BD-LM, we compare three configurations:
(1)~\textbf{SingleBD (Native)}, the model's native single-block diffusion inference;
(2)~\textbf{MultiBD (training-free)}, MultiBD inference applied without post-training; and
(3)~\textbf{MBD-*}, the corresponding MultiTF-post-trained model using MultiBD inference.

\paragraph{Main analysis.}
MBD-LMs improve decoding parallelism while preserving generation quality.
Compared with LLaDA2-Mini under SingleBD (Native), MBD-LLaDA2-Mini increases average TPF from 3.47 to \textbf{6.19} (+78.4\%) and improves average accuracy from 79.95\% to \textbf{81.03\%}.
Notably, even without DMax, MBD-LLaDA2-Mini reaches a TPF comparable to LLaDA2-Mini-DMax under SingleBD (6.19 vs. 6.35), while achieving higher average accuracy (81.03\% vs. 79.59\%).
This shows that MultiTF can turn a standard BD-LM into an MBD-LM with DMax-level decoding parallelism.

\paragraph{Compatibility with T2T-enhanced decoding.}
MultiTF is complementary to DMax, a Token-to-Token (T2T) enhanced acceleration method.
When combined with DMax, MBD-LLaDA2-Mini-DMax further increases average TPF from 6.35 to \textbf{9.34} (+47.1\%) over LLaDA2-Mini-DMax under SingleBD, with only a 1.02 percentage-point average accuracy drop.
This indicates that MBD-LMs can stack with existing T2T-enhanced recipes.

\paragraph{Effect of train--inference alignment.}
The comparison between training-free MultiBD and MultiTF-post-trained MBD-LMs highlights the importance of train--inference alignment.
Directly applying MultiBD already increases TPF, confirming that multi-block decoding relaxes the single-block bottleneck.
However, it can degrade accuracy because the original BD-LMs are not trained on practical MultiBD states.
MultiTF reduces this mismatch: on LLaDA2-Mini, accuracy improves from 78.59\% under training-free MultiBD to \textbf{81.03\%} after MultiTF post-training, while average TPF further increases from 4.41 to \textbf{6.19}.
On LLaDA2-Mini-DMax, MultiTF improves average accuracy from 78.17\% to \textbf{78.57\%} while preserving high TPF.

\paragraph{Generalization across BD-LM backbones.}
MBD-LMs also generalize beyond the LLaDA2 family.
On SDAR-8B-Chat-b32, MBD-SDAR-8B-Chat-b32 increases average TPF from 2.54 to \textbf{4.46} (+75.6\%) and improves average accuracy from 69.00\% to \textbf{69.74\%}.
With block size 4, MBD-SDAR-8B-Chat-b4 reaches the best average AUP among the three SDAR configurations.
These results suggest that the MBD-LM formulation and MultiTF post-training are not tied to a specific BD-LM backbone.

\paragraph{Transfer of training-free MultiBD.}
In addition, Table~\ref{tab:transfer_math} shows that training-free MultiBD transfers to additional model variants such as LLaDA2-Mini-CAP and LLaDA2.1-Mini, improving TPF without post-training.
This suggests that the inference-side MultiBD mechanism itself has broad applicability, while MultiTF is needed to recover and further improve generation quality under practical MultiBD states.

\subsection{Ablation Study}

Table~\ref{tab:ablation} ablates the key MultiTF training components with LLaDA2-Mini-DMax, averaged over HumanEval+ and GSM8K.
Compared with SingleBD (Native), the full MBD configuration increases TPF from 6.57 to \textbf{9.87} and AUP from 536.89 to \textbf{805.34}, while nearly preserving the average accuracy, with only a 0.08-point change from 84.67\% to 84.59\%.
This shows that MultiTF substantially improves the TPF--accuracy trade-off by aligning BD-LMs with practical MultiBD inference states.

\paragraph{Effect of \textit{noise-group} \textit{group-layouts}.}
We first ablate the \textit{group-layout} construction for \textit{noise-groups}.
Using only systematic layouts or only random layouts already improves TPF over SingleBD, increasing TPF from 6.57 to 9.71 and 9.42, respectively.
However, both single-source variants reduce accuracy, with systematic layouts achieving 83.22\% and random layouts achieving 82.72\%.
Combining systematic and random layouts gives the best trade-off, reaching the highest TPF of \textbf{9.87} and the highest AUP of \textbf{805.34}, while recovering the accuracy to 84.59\%, close to the SingleBD level of 84.67\%.
This suggests that the two layout sources are complementary:
systematic \textit{group-layouts} provide structured coverage of bounded \textit{running-set} sizes and group-relative positions, while random \textit{group-layouts} add distributional diversity beyond the systematic construction.

\paragraph{Effect of block-level \textit{noise-schedulers}.}
We then ablate the block-level \textit{noise-scheduler} within each \textit{noise-group}.
Replacing the chain-uniform \textit{noise-scheduler} with a D2F-style monotonic \textit{noise-scheduler} increases TPF over SingleBD from 6.57 to 8.76, but causes a large accuracy drop from 84.67\% to 79.34\%.
This indicates that exposing the model to multiple noisy blocks is insufficient when the slot-wise noise pattern is not aligned with practical MultiBD inference.
Random and sorted-uniform \textit{noise-schedulers} further improve TPF to 9.70 and 9.73, respectively, but still underperform chain-uniform in AUP.
In particular, sorted-uniform achieves a high TPF but suffers a larger accuracy drop, suggesting that sorted mask ratios alone do not capture the heterogeneous noise gaps induced by MultiBD decoding.
The full chain-uniform \textit{noise-scheduler} achieves the best accuracy, TPF, and AUP among the scheduler variants, reaching 84.59\%, 9.87, and 805.34, respectively.
This confirms the importance of training with heterogeneous slot-wise noise gaps.
The sorted-uniform \textit{noise-scheduler} baseline samples mask ratios uniformly and sorts them before assigning them to slots; details are provided in Appendix~\ref{sec:appendix_training}.
We further analyze the train--inference alignment gap in Section~\ref{sec:train_inference_alignment}.

\subsection{Train--Inference Alignment Analysis}
\label{sec:train_inference_alignment}

Figure~\ref{fig:train_infer_noise_activity} analyzes the training-state mismatch that motivates MultiTF.
The figure focuses on two aspects of practical MultiBD inference:
slot-wise mask-ratio patterns and the size of the active part of the \textit{running-set}.

\paragraph{D2F-style noise schedules mismatch MultiBD inference.}
As shown in Figure~\ref{fig:train_infer_noise_activity}(A), the D2F-style monotonic scheduler induces highly overlapping slot-wise mask-ratio distributions.
This weak slot-wise separation differs from practical MultiBD inference, where adjacent active slots often exhibit large noise-ratio gaps.
This explains the ablation result in Table~\ref{tab:ablation}: the D2F-style monotonic \textit{noise-scheduler} improves TPF by enabling multi-block decoding, but causes a large accuracy drop because its training states do not match practical MultiBD inference states.

\paragraph{Chain-uniform scheduling improves slot-wise alignment.}
By contrast, the chain-uniform scheduler used by MultiTF creates more heterogeneous slot-wise noise patterns.
As shown in Figure~\ref{fig:train_infer_noise_activity}(B), different slots in a \textit{noise-group} receive more separated mask-ratio distributions.
These scheduler-induced training distributions better match the inference-time mask-ratio distributions in Figure~\ref{fig:train_infer_noise_activity}(C), especially the large gap between the first and second active slots.
After MultiTF post-training, the inference-time mask-ratio distribution becomes further aligned with the designed training states.

\paragraph{MultiBD inference uses a bounded active set.}
Figure~\ref{fig:train_infer_noise_activity}(D--E) further shows that MultiBD inference usually maintains a small active part of the \textit{running-set}, with an expectation around two and occasional expansion to three or four active blocks.
This supports the bounded \textit{running-set} view in Section~\ref{sec:mbd_lms}.
Reliable MultiBD therefore requires training states that match both the bounded \textit{running-set} structure and the heterogeneous slot-wise noise patterns of inference, rather than merely exposing the model to future noisy blocks.

\subsection{Efficiency Analysis}
\label{sec:efficiency_analysis}

We further analyze how the increased TPF of MBD-LMs translates into realized wall-clock throughput.
At decoding step $s$, the optimized MultiBD engine executes a fixed physical \textit{Block Buffer} $\mathcal{W}_s$ defined in Section~\ref{sec:decoding}.
Let $P_s$ denote the cached prefix length at this step and let
\[
Q_s = |\mathcal{W}_s|B = N_{\mathrm{buf}}B
\]
denote the number of processed tokens in one forward pass.
For SingleBD, this reduces to $N_{\mathrm{buf}}=1$ and $Q_s=B$.
For MultiBD, $N_{\mathrm{buf}}>1$, and the forward pass processes all physical buffer slots, including active blocks, completed resident blocks, and dummy slots used to preserve static input shapes.
Thus, $Q_s$ measures the computational workload of a forward pass, whereas TPF measures the number of useful tokens committed by that forward pass.

This distinction defines a token-efficiency factor:
\[
\eta_{\mathrm{tok}}(s)
=
\frac{\mathrm{TPF}_s}{Q_s}.
\]
Equivalently,
\[
\mathrm{TPS}
=
\frac{\mathrm{TPF}}{T_{\mathrm{step}}}
=
\frac{\eta_{\mathrm{tok}} Q_s}{T_{\mathrm{step}}}.
\]
Therefore, increasing the block-buffer size can improve throughput only when the useful-token gain outweighs the additional per-step cost.
MultiBD increases $Q_s$ and enables more tokens to be committed per forward pass, but its token efficiency can be reduced by inactive dummy slots and resident blocks that are processed for static-shape execution but do not immediately contribute to committed tokens.

Each decoding forward can be viewed as an extend-attention step with $Q_s$ query tokens and a cached prefix of length $P_s$.
For a transformer with $N_{\mathrm{layer}}$ layers, hidden size $d$, FFN hidden size $d_{\mathrm{ff}}$, and vocabulary $\mathcal{V}$, the per-step FLOPs can be approximated as
\[
\mathcal{F}_{\mathrm{step}}(Q_s,P_s)
=
\Theta\left(
N_{\mathrm{layer}}
\left[
Q_s(d^2+d d_{\mathrm{ff}})
+
d(Q_sP_s+Q_s^2)
\right]
+
Q_s d |\mathcal{V}|
\right).
\]
The first term comes from QKV/O projections and FFN layers, the second term comes from attention between the buffer and the cached prefix as well as attention inside the buffer, and the last term comes from the LM head when logits are computed.
Thus, increasing $N_{\mathrm{buf}}$ from $1$ to a larger value improves inter-block decoding parallelism, but also increases the amount of computation performed by each forward pass.

The memory cost follows the same extend-attention structure.
Let $s_{\mathrm{dtype}}$ be the number of bytes per activation element.
The per-step weight traffic scales as
\[
\mathcal{M}_{W}
=
\Theta\left(
N_{\mathrm{layer}}s_{\mathrm{dtype}}(d^2+d d_{\mathrm{ff}})
\right),
\]
while the KV-cache traffic of extend attention can be approximated as
\[
\mathcal{M}_{\mathrm{KV}}(Q_s,P_s)
=
\Theta\left(
N_{\mathrm{layer}}s_{\mathrm{dtype}}
\left[
\rho(Q_s)(P_s+Q_s)d + Q_s d
\right]
\right),
\]
where $\rho(Q_s)$ captures repeated KV reads caused by query tiling.
The first term corresponds to reading KV cache for the prefix and current buffer, while the second term corresponds to KV cache storing.

This gives a roofline-style view of the step latency:
\[
T_{\mathrm{step}}(Q_s,P_s)
\approx
\max\left(
\frac{\mathcal{F}_{\mathrm{step}}(Q_s,P_s)}{\Pi_{\mathrm{eff}}},
\frac{\mathcal{M}_{W}+\mathcal{M}_{\mathrm{KV}}(Q_s,P_s)}{\mathcal{B}_{\mathrm{HBM}}}
\right)
+
T_{\mathrm{comm}}(Q_s)
+
T_{\mathrm{launch}},
\]
where $\Pi_{\mathrm{eff}}$ is the effective compute throughput, $\mathcal{B}_{\mathrm{HBM}}$ is the effective HBM bandwidth, $T_{\mathrm{comm}}$ includes fixed-configuration tensor-parallel communication, and $T_{\mathrm{launch}}$ denotes launch and runtime overhead.
This expression shows that the realized throughput depends on both the useful-token numerator and the roofline-limited per-step cost denominator.

The attention arithmetic intensity further explains why MultiBD can still be efficient despite processing more tokens per step.
Ignoring lower-order terms, the attention arithmetic intensity is approximately
\[
\mathrm{AI}_{\mathrm{attn}}
\approx
\frac{dQ_sP_s}{s_{\mathrm{dtype}}\rho(Q_s)P_s d}
=
\Theta\left(
\frac{Q_s}{s_{\mathrm{dtype}}\rho(Q_s)}
\right)
\]
when $P_s \gg Q_s$.
Therefore, increasing $Q_s$ through a larger \textit{Block Buffer} makes the extend-attention step more compute intensive.
Prefix KV reads, weight reads, and kernel-launch overheads are amortized over more query tokens.
However, the gain is useful only to the extent that these processed tokens lead to committed tokens, as captured by $\eta_{\mathrm{tok}}$.

The measurements in Table~\ref{tab:tps_comparison} match this analysis.
For LLaDA2-Mini, MBD increases the average TPF from 3.47 to 6.19, a $1.78\times$ improvement, while the step latency increases from 7.07 ms to 8.78 ms, a $1.24\times$ cost increase.
The expected throughput scaling is therefore approximately $1.78/1.24=1.44\times$, closely matching the measured Avg. TPS improvement from 517.16 to 745.92, i.e., $1.44\times$.
Similarly, for LLaDA2-Mini-DMax, MBD increases the average TPF from 6.35 to 9.34, a $1.47\times$ improvement, while the step latency increases from 9.02 ms to 11.20 ms, a $1.24\times$ cost increase.
This predicts a throughput scaling of $1.47/1.24=1.18\times$, which closely matches the measured Avg. TPS improvement from 779.49 to 926.67, i.e., $1.19\times$.
Thus, the observed gap between TPF gain and TPS gain is primarily explained by the increased per-forward cost of processing the larger static \textit{Block Buffer}.

Overall, MultiBD improves wall-clock throughput by increasing the number of useful tokens committed per forward pass and by making each extend-attention step more compute intensive.
At the same time, static-shape execution introduces extra processed tokens through resident blocks and dummy slots, reducing token efficiency relative to the ideal case.
The final TPS gain is therefore determined by the balance among TPF improvement, token efficiency, and roofline-limited step latency.


\begin{table*}[t]
\centering
\caption{
Throughput and single-step latency comparison.
Results are measured for single-sample decoding on two H100 GPUs with tensor parallelism degree 2 (TP=2).
Step latency denotes the average wall-clock latency of one decoding forward pass.
TPF and TPS gains are computed relative to LLaDA2-Mini, while latency cost reports the relative increase in per-step latency.
}
\label{tab:tps_comparison}
\setlength{\tabcolsep}{3.2pt}
\renewcommand{\arraystretch}{1.08}
\resizebox{\textwidth}{!}{%
\begin{tabular}{lcccccccccc}
\toprule
& \multicolumn{4}{c}{\textbf{Forward-step statistics}} & \multicolumn{6}{c}{\textbf{Realized throughput}} \\
\cmidrule(lr){2-5} \cmidrule(lr){6-11}
\textbf{Model} 
& \textbf{Avg. TPF $\uparrow$} 
& \textbf{TPF Gain $\uparrow$} 
& \textbf{Step Lat. (ms) $\downarrow$} 
& \textbf{Lat. Cost $\downarrow$} 
& \textbf{GSM8K TPS $\uparrow$} 
& \textbf{MATH500 TPS $\uparrow$} 
& \textbf{MBPP+ TPS $\uparrow$} 
& \textbf{HumanEval+ TPS $\uparrow$} 
& \textbf{Avg. TPS $\uparrow$} 
& \textbf{TPS Gain $\uparrow$} \\
\midrule
LLaDA2-Mini 
& 3.47 & -- & \textbf{7.07} & $1.00\times$ 
& 344.05 & 403.45 & 496.19 & 824.94 & 517.16 & -- \\
\midrule
MBD-LLaDA2-Mini 
& 6.19 & +78.39\% & 8.78 & $1.24\times$ 
& 687.87 & 707.89 & 646.73 & 941.18 & 745.92 & +44.24\% \\
LLaDA2-Mini-DMax 
& 6.35 & +83.00\% & 9.02 & $1.28\times$ 
& 700.82 & 730.60 & 754.97 & 931.55 & 779.49 & +50.73\% \\
MBD-LLaDA2-Mini-DMax 
& \textbf{9.34} & \textbf{+169.16\%} & 11.20 & $1.58\times$ 
& \textbf{834.52} & \textbf{851.07} & \textbf{896.65} & \textbf{1124.43} & \textbf{926.67} & \textbf{+79.19\%} \\
\bottomrule
\end{tabular}%
}
\end{table*}




\FloatBarrier

%% file: sections/06_related_work.tex
\section{Related Work}

\subsection{Diffusion Language Models}

Diffusion Language Models (DLMs) generate text through iterative denoising and enable parallel token refinement as an alternative to autoregressive generation.
Representative models include LLaDA~\citep{nie2025llada}, Dream~\citep{ye2025dream}, and LLaDA2.x~\citep{bie2025llada2,bie2026llada21}, which improve scaling, initialization, and editable refinement.
However, fully bidirectional DLMs are difficult to serve efficiently because they do not naturally support KV caching or flexible-length generation.

Block Diffusion Language Models (BD-LMs)~\citep{arriola2025block,bie2025llada2,cheng2025sdar} address these limitations by introducing block-causal generation.
Their native \textit{Single-Block Diffusion} (SingleBD) inference decodes one noisy block conditioned on a clean cached prefix, enabling KV caching and intra-block parallel decoding.
Nevertheless, SingleBD still processes blocks sequentially, leaving inter-block parallelism underused.
Our work studies \textit{Multi-Block Diffusion} (MultiBD) as a broader inference regime for BD-LMs, where a bounded \textit{running-set} of consecutive blocks can be refined concurrently.

\subsection{Efficient DLM Inference and Training}

Efficient DLMs have been studied through distillation, scheduling, caching, and parallel decoding.
D2F~\citep{wang2025d2f} introduces noisy-block visibility during training and demonstrates the potential of MultiBD-style pipelined decoding.
DMax~\citep{chen2026dmax}, d3LLM~\citep{qian2026d3llm}, LightningRL~\citep{hu2026lightningrl}, and dParallel~\citep{chen2025dparallel} improve the accuracy--parallelism trade-off through training objectives or decoding schedules.
Fast-dLLM~\citep{wu2025fastdllm} and LoPA~\citep{xu2025lopa} accelerate inference through caching and lookahead parallelism.

Our work is complementary to these efforts but focuses on a different level of parallelism.
Instead of only increasing token-level parallelism or applying MultiBD as an inference-time heuristic, we treat MultiBD as a target inference regime for BD-LMs.
We identify the bounded \textit{running-set} structure and heterogeneous slot-wise noise patterns as key train--inference alignment factors, and propose MultiTF to post-train BD-LMs into MBD-LMs with inference-like multi-block states.
We further provide \textit{Block Buffer} inference support so that MultiBD preserves prefix-cache reuse and static-shape execution.

%% file: sections/07_conclusion.tex
\section{Conclusion}

We proposed \textit{Multi-Block Diffusion Language Models} (MBD-LMs), a unified formulation of BD-LMs for reliable MultiBD inference.
Starting from the sequential bottleneck of SingleBD, we showed that MultiBD can expose inter-block parallelism but requires training states aligned with its bounded \textit{running-set} structure and heterogeneous slot-wise noise patterns.
To bridge this gap, we introduced \textit{Multi-block Teacher Forcing} (MultiTF), which post-trains BD-LMs with bounded \textit{noise-groups}, the \textit{Group-Aware Dual-Stream Mask}, and randomized block-level \textit{noise-schedulers}.
We further developed an optimized MultiBD inference engine with the \textit{Block Buffer} mechanism, enabling static-shape execution while preserving KV caching and prefix-cache reuse.
Experiments on math and code benchmarks show that MBD-LMs improve decoding parallelism and realized throughput while maintaining generation quality, demonstrating that reliable MultiBD requires both training-time state alignment and inference-time system support.

%% file: sections/08_appendix_proof.tex
\section{Theoretical View of MultiTF}
\label{sec:appendix_theory}

This appendix provides a simple theoretical view of Multi-block Teacher Forcing (MultiTF).
The goal is not to prove that MultiTF directly improves downstream accuracy.
Instead, we show that MultiTF can be interpreted as a coverage-based surrogate for the ideal MultiBD training objective, and that its approximation gap is controlled by the mismatch in \textit{running-set} coverage and noise-ratio distributions.

\paragraph{Ideal MultiBD objective.}
Let $\mathcal{R}=\{a,\ldots,c\}$ denote a consecutive MultiBD \textit{running-set} with size $|\mathcal{R}|\le G_{\max}$, where $G_{\max}$ is the maximum \textit{noise-group} size used in MultiTF.
For a clean sequence $\mathbf{x}_0$, noise ratios $\mathbf{t}$, and model $\theta$, define the state loss
\begin{equation}
\begin{aligned}
\ell_\theta(\mathbf{x}_0,\mathcal{R},\mathbf{t})
=
-\frac{1}{|\mathcal{M}_{\mathcal{R}}|}
\sum_{i\in\mathcal{M}_{\mathcal{R}}}
\log p_\theta
\bigl(
x_0^i
\mid
\mathbf{x}_0^{(<a)},
\mathbf{b}_{\mathcal{R},\mathbf{t}}
\bigr),
\end{aligned}
\label{eq:state_loss_theory}
\end{equation}
where $\mathbf{x}_0^{(<a)}$ is the clean prefix before $\mathcal{R}$, $\mathbf{b}_{\mathcal{R},\mathbf{t}}$ denotes the noisy blocks inside $\mathcal{R}$, and $\mathcal{M}_{\mathcal{R}}$ denotes masked positions in the \textit{running-set}.

Let $p_{\mathrm{inf}}(\mathcal{R},\mathbf{t})$ be the inference-time distribution of MultiBD states, and let $q_{\mathrm{MultiTF}}(\mathcal{R},\mathbf{t})$ be the training-state distribution induced by MultiTF \textit{group-layouts} and the chain-uniform \textit{noise-scheduler}.
The ideal MultiBD objective is
\begin{equation}
\mathcal{L}_{\mathrm{MultiBD}}^\star(\theta)
=
\mathbb{E}_{\mathbf{x}_0}
\mathbb{E}_{(\mathcal{R},\mathbf{t})\sim p_{\mathrm{inf}}}
\left[
\ell_\theta(\mathbf{x}_0,\mathcal{R},\mathbf{t})
\right],
\label{eq:ideal_mbd_objective}
\end{equation}
while MultiTF minimizes the surrogate objective
\begin{equation}
\mathcal{L}_{\mathrm{MultiTF}}(\theta)
=
\mathbb{E}_{\mathbf{x}_0}
\mathbb{E}_{(\mathcal{R},\mathbf{t})\sim q_{\mathrm{MultiTF}}}
\left[
\ell_\theta(\mathbf{x}_0,\mathcal{R},\mathbf{t})
\right].
\label{eq:multitf_surrogate_objective}
\end{equation}

\paragraph{Systematic shifts cover bounded \textit{running-sets}.}
Assume the sequence is padded so that boundary effects can be ignored.
For a fixed \textit{noise-group} size $g\in\{2,\ldots,G_{\max}\}$, MultiTF constructs $g$ shifted layouts.
Then every consecutive \textit{running-set} $\mathcal{R}=\{a,\ldots,a+g-1\}$ appears as one \textit{noise-group} in exactly one shifted layout for that $g$.

\noindent\textit{Proof.}
For a fixed $g$, each shifted layout places group boundaries every $g$ blocks with a different offset.
For a \textit{running-set} starting at block $a$, choosing the shift $h=(a-1)\bmod g$ aligns a group boundary with $a$, so $\{a,\ldots,a+g-1\}$ appears as one \textit{noise-group}.
The shift is unique modulo $g$, so the \textit{running-set} appears once among the $g$ shifted layouts.

Thus, systematic layouts cover all consecutive \textit{running-sets} with size between $2$ and $G_{\max}$.
Equivalently, for each fixed $g$, every block appears once at every group-relative logical slot across the $g$ shifts.
Random layouts do not change this support guarantee, but add additional samples with non-regular \textit{noise-group}-size combinations.

\paragraph{Objective mismatch bound.}
We next bound the gap between the ideal MultiBD objective and the MultiTF surrogate objective.
Let $p_{\mathcal{R}}$ and $q_{\mathcal{R}}$ be the marginal distributions over \textit{running-sets} under $p_{\mathrm{inf}}$ and $q_{\mathrm{MultiTF}}$, respectively.

We assume:

\textbf{A1. Bounded MultiBD states.}
The inference distribution $p_{\mathrm{inf}}$ is supported on consecutive \textit{running-sets} with $2\le|\mathcal{R}|\le G_{\max}$.

\textbf{A2. Bounded loss.}
For all $\theta,\mathbf{x}_0,\mathcal{R},\mathbf{t}$,
\[
0\le \ell_\theta(\mathbf{x}_0,\mathcal{R},\mathbf{t})\le M .
\]

\textbf{A3. Lipschitz dependence on noise ratios.}
For every $\theta,\mathbf{x}_0,\mathcal{R}$, the state loss is $L_t$-Lipschitz in the noise-ratio vector:
\begin{equation}
\left|
\ell_\theta(\mathbf{x}_0,\mathcal{R},\mathbf{t})
-
\ell_\theta(\mathbf{x}_0,\mathcal{R},\mathbf{t}')
\right|
\le
L_t\|\mathbf{t}-\mathbf{t}'\|_1 .
\label{eq:lipschitz_noise_assumption}
\end{equation}

Here $\mathrm{TV}(p,q)=\frac{1}{2}\sum_x |p(x)-q(x)|$ denotes the total variation distance between two discrete distributions.

Define the \textit{running-set} distribution mismatch as
\[
\delta_{\mathcal{R}}
=
\mathrm{TV}(p_{\mathcal{R}},q_{\mathcal{R}}),
\]
and assume the conditional noise-ratio mismatch satisfies
\[
W_1
\left(
p_{\mathrm{inf}}(\mathbf{t}\mid\mathcal{R}),
q_{\mathrm{MultiTF}}(\mathbf{t}\mid\mathcal{R})
\right)
\le
\delta_t
\]
for every \textit{running-set} $\mathcal{R}$, where $W_1$ is the Wasserstein-1 distance under the $\ell_1$ metric.

Under these assumptions, for any model $\theta$,
\begin{equation}
\left|
\mathcal{L}_{\mathrm{MultiBD}}^\star(\theta)
-
\mathcal{L}_{\mathrm{MultiTF}}(\theta)
\right|
\le
M\delta_{\mathcal{R}} + L_t\delta_t .
\label{eq:objective_mismatch_bound}
\end{equation}

\noindent\textit{Proof.}
For clarity, omit the outer expectation over $\mathbf{x}_0$.
We decompose the objective gap into a \textit{running-set} distribution term and a conditional noise-distribution term:
\begin{equation}
\begin{aligned}
\left|
\mathbb{E}_{p_{\mathcal{R}}p(\mathbf{t}\mid\mathcal{R})}
[\ell_\theta]
-
\mathbb{E}_{q_{\mathcal{R}}q(\mathbf{t}\mid\mathcal{R})}
[\ell_\theta]
\right|
\le
\left|
\mathbb{E}_{p_{\mathcal{R}}p(\mathbf{t}\mid\mathcal{R})}
[\ell_\theta]
-
\mathbb{E}_{q_{\mathcal{R}}p(\mathbf{t}\mid\mathcal{R})}
[\ell_\theta]
\right|
+
\left|
\mathbb{E}_{q_{\mathcal{R}}p(\mathbf{t}\mid\mathcal{R})}
[\ell_\theta]
-
\mathbb{E}_{q_{\mathcal{R}}q(\mathbf{t}\mid\mathcal{R})}
[\ell_\theta]
\right|.
\end{aligned}
\end{equation}
The first term is bounded by $M\mathrm{TV}(p_{\mathcal{R}},q_{\mathcal{R}})=M\delta_{\mathcal{R}}$, since the loss is bounded in $[0,M]$.
The second term is bounded by $L_t\delta_t$ by the Lipschitz assumption and the definition of $W_1$.
Combining the two terms gives Eq.~\ref{eq:objective_mismatch_bound}.

\paragraph{Excess target risk.}
Let $\hat{\theta}$ be a model whose MultiTF objective is within $\epsilon_{\mathrm{opt}}$ of the best model in a hypothesis class $\Theta$:
\[
\mathcal{L}_{\mathrm{MultiTF}}(\hat{\theta})
\le
\min_{\theta\in\Theta}\mathcal{L}_{\mathrm{MultiTF}}(\theta)
+
\epsilon_{\mathrm{opt}}.
\]
Then
\begin{equation}
\mathcal{L}_{\mathrm{MultiBD}}^\star(\hat{\theta})
-
\min_{\theta\in\Theta}\mathcal{L}_{\mathrm{MultiBD}}^\star(\theta)
\le
2(M\delta_{\mathcal{R}}+L_t\delta_t)
+
\epsilon_{\mathrm{opt}} .
\label{eq:excess_target_risk}
\end{equation}

This bound shows that reducing \textit{running-set} distribution mismatch $\delta_{\mathcal{R}}$ and noise-ratio mismatch $\delta_t$ directly tightens the gap between MultiTF training and ideal MultiBD inference.
Systematic shifts reduce support mismatch by covering bounded consecutive \textit{running-sets} up to size $G_{\max}$, random layouts add distributional diversity, and the chain-uniform \textit{noise-scheduler} reduces noise-ratio mismatch by producing heterogeneous slot-wise noise gaps.
Therefore, MultiTF can be viewed as a coverage-based surrogate for the ideal MBD-LM objective.

%% file: sections/09_appendix_training_implementation.tex
\section{MultiTF Training Implementation Details}
\label{sec:appendix_training}

This appendix provides implementation details for Multi-block Teacher Forcing (MultiTF), which post-trains BD-LMs into MBD-LMs.
The terminology follows Section~\ref{sec:training}: training-side structures are called \textit{noise-groups}, \textit{group-layouts}, and \textit{noise-schedulers}, while inference-side structures are called \textit{Block Buffers} and \textit{slots}.
We use $G_{\max}$ for the maximum \textit{noise-group} size, $\Lambda$ for the set of \textit{group-layouts}, $\lambda$ for one \textit{group-layout}, and $H_m$ for one \textit{noise-group}.
We use VeOmni~\citep{ma2025veomni} as the training framework.
SDAR models are post-trained on reasoning/code data from prior studies~\citep{boizard2025reasoning,jtatman2025pythoncode500k}; LLaDA2.x and DMax-enhanced models are post-trained on the corresponding reasoning/code mixtures used by their base recipes.

\subsection{Group-Layout Construction}
\label{sec:appendix_layout_construction}

MultiTF constructs a \textit{group-layout} set
\[
\Lambda=\Lambda_{\mathrm{sys}}\cup\Lambda_{\mathrm{rand}},
\]
where $\Lambda_{\mathrm{sys}}$ contains systematic shifted layouts and $\Lambda_{\mathrm{rand}}$ contains random layouts.
Each \textit{group-layout} $\lambda=(H_1,\ldots,H_{|\lambda|})$ partitions the block sequence $[\mathbf{b}_1,\ldots,\mathbf{b}_K]$ into consecutive \textit{noise-groups}.
Each \textit{noise-group} $H_m=\{a_m,\ldots,c_m\}$ has the same consecutive-block form as a possible MultiBD \textit{running-set}.

\paragraph{Systematic layouts.}
For each \textit{noise-group} size $g\in\{2,\ldots,G_{\max}\}$ and shift $h\in\{0,\ldots,g-1\}$, MultiTF constructs a shifted layout $\lambda_{g,h}$ by placing group boundaries every $g$ blocks with offset $h$.
Formally, define the boundary set
\[
\mathcal{B}_{g,h}
=
\mathrm{sort}
\Big(
\{1,K+1\}
\cup
\{\,1+h+qg: q\in\mathbb{Z},\ 1<1+h+qg<K+1\,\}
\Big).
\]
Let $\mathcal{B}_{g,h}=(r_1,\ldots,r_{n_{g,h}+1})$ after sorting.
The $q$-th \textit{noise-group} in $\lambda_{g,h}$ is
\[
H_{g,h,q}
=
\{r_q,\ldots,r_{q+1}-1\},
\qquad
q=1,\ldots,n_{g,h}.
\]
Boundary \textit{noise-groups} can be shorter than $g$, while interior \textit{noise-groups} have size $g$.
The systematic layout set is
\[
\Lambda_{\mathrm{sys}}
=
\{\lambda_{g,h}: g\in\{2,\ldots,G_{\max}\},\ h\in\{0,\ldots,g-1\}\}.
\]
Ignoring boundary effects, every consecutive \textit{running-set} $\{a,\ldots,a+g-1\}$ of length $g$ appears as one \textit{noise-group} in exactly one shifted layout by choosing $h=(a-1)\bmod g$.
Equivalently, for each fixed $g$, every block appears once at every group-relative position across the $g$ shifts.
The number of systematic layouts is therefore
\begin{equation}
|\Lambda_{\mathrm{sys}}|
=
\sum_{g=2}^{G_{\max}} g
=
\frac{(G_{\max}+2)(G_{\max}-1)}{2}.
\end{equation}

\paragraph{Random layouts.}
Systematic layouts provide structured coverage but are regular by construction.
To increase layout diversity, MultiTF further samples $N_{\mathrm{rand}}$ random layouts.
For each random layout, we sequentially draw group sizes
\[
g_m\sim \mathrm{Uniform}\{2,\ldots,G_{\max}\}
\]
and form consecutive groups
\[
H_m=\{a_m,\ldots,\min(a_m+g_m-1,K)\},
\qquad
a_{m+1}=\min(a_m+g_m,K+1),
\]
until the full block sequence is covered.
These random layouts add non-regular \textit{noise-group}-size combinations and boundary patterns without replacing the coverage guarantee of systematic layouts.
The total number of layout variants per clean sequence is
\begin{equation}
|\Lambda|
=
\frac{(G_{\max}+2)(G_{\max}-1)}{2}
+
N_{\mathrm{rand}}.
\end{equation}
All layouts are batched as independent input sequences during post-training.
This increases the effective number of training states per clean sample, but also increases training cost; exact settings are reported in Table~\ref{tab:appendix_train_hparams}.
A theoretical coverage view is provided in Appendix~\ref{sec:appendix_theory}.

\subsection{Chain-uniform Noise-Scheduler}
\label{sec:appendix_chain_uniform_scheduler}

For each \textit{noise-group} $H_m=(j_1,\ldots,j_{n_m})$, MultiTF applies the chain-uniform \textit{noise-scheduler} used in Algorithm~\ref{alg:multitf_training}.
We first define an effective upper bound
\begin{equation}
t_{\mathrm{eff}}
=
t_{\mathrm{high}}
-
\rho(t_{\mathrm{high}}-t_{\mathrm{low}}),
\end{equation}
where $\rho$ is the noise-transition margin ratio, corresponding to \texttt{noise\_transition\_margin\_ratio} in the implementation.
This parameter is independent of the random \textit{noise-scheduler} power-law bias $\gamma_{\mathrm{rand}}$, which is used only for the random \textit{noise-scheduler} ablation.

For each group, a group-level floor $\ell$ is first sampled from the lower part of the noise range.
Then each block samples its mask ratio from the interval between the current floor and the effective upper bound, and the sampled ratio becomes the floor for the next block:
\begin{equation}
\begin{aligned}
\ell \sim \mathcal{U}(t_{\mathrm{low}},t_{\mathrm{eff}}),\quad 
t_{j_i} \sim \mathcal{U}(\ell,t_{\mathrm{eff}}),\quad 
\ell &\leftarrow t_{j_i},
\quad i=1,\ldots,n_m.
\end{aligned}
\end{equation}
This construction produces monotonic but randomized slot-wise mask ratios inside each \textit{noise-group}.
Compared with the fixed-step D2F schedule over a long noisy sequence, the resulting groups have larger and more variable block-level noise-ratio gaps, matching the heterogeneous active blocks observed during MultiBD inference.

For each block with mask ratio $t_{j_i}$, MultiTF replaces $\lfloor B\cdot t_{j_i}\rfloor$ randomly selected token positions in $\mathbf{b}_{j_i}$ with \texttt{[M]}.
For a layout $\lambda$, the resulting noisy sequence is denoted as $\mathbf{x}^{\lambda}_{\mathbf{t}}$.

\subsection{Group-Aware Dual-Stream Mask}
\label{sec:appendix_group_aware_mask}

Following the TF-style construction of Block Diffusion, MultiTF concatenates the noisy and clean sequences into the input sequence
\begin{equation}
\mathbf{X}_{\lambda}
=
[\mathbf{x}^{\lambda}_{\mathbf{t}};\mathbf{x}_0].
\end{equation}
The attention mask has the block form
\begin{equation}
\mathbf{A}_{\lambda}
=
\begin{bmatrix}
\mathbf{M}_{\mathrm{GD}} & \mathbf{M}_{\mathrm{GOC}} \\
0 & \mathbf{M}_{\mathrm{BC}}
\end{bmatrix},
\end{equation}
where $\mathbf{M}_{\mathrm{GD}}$ is the group-aware diagonal mask on the noisy part, $\mathbf{M}_{\mathrm{GOC}}$ is the group-aware offset-causal mask from noisy tokens to clean tokens, and $\mathbf{M}_{\mathrm{BC}}$ is the standard block-causal mask on the clean part.

Let $\mathcal{N}_{\lambda}$ and $\mathcal{C}$ denote token positions in the noisy and clean parts, respectively.
Let $g(i)$ be the \textit{noise-group} index of token $i$, $\beta(i)$ be its block index, and $\alpha(i)$ be the first block index of the \textit{noise-group} containing $i$.
The three masks are defined as
\begin{align}
[\mathbf{M}_{\mathrm{GD}}]_{ij}=1
&\iff
i,j\in\mathcal{N}_{\lambda},
\quad
g(i)=g(j),
\quad
\beta(j)\le\beta(i),
\\
[\mathbf{M}_{\mathrm{GOC}}]_{ij}=1
&\iff
i\in\mathcal{N}_{\lambda},
\quad
j\in\mathcal{C},
\quad
\beta(j)<\alpha(i),
\\
[\mathbf{M}_{\mathrm{BC}}]_{ij}=1
&\iff
i,j\in\mathcal{C},
\quad
\beta(j)\le\beta(i),
\end{align}
and all other entries are zero.
Thus, noisy tokens can attend to same-\textit{noise-group} noisy tokens from the same or preceding blocks, each \textit{noise-group} can condition on clean prefix blocks before it, and clean tokens never attend to noisy tokens.
This implements the visibility pattern required by Equation~\ref{eq:multibd_state} without information leakage.



\subsection{MultiTF Objective and Model-specific Training Recipes}
\label{sec:appendix_objectives}

MultiTF defines the training-state construction: the layout $\lambda$, the noisy sequence $\mathbf{x}^{\lambda}_{\mathbf{t}}$, the clean sequence $\mathbf{x}_0$, and the \textit{Group-Aware Dual-Stream Mask} $\mathbf{A}_{\lambda}$.
Different base BD-LMs can reuse the same MultiTF input sequences while keeping their own model-specific training recipes.

\subsubsection{Default MultiTF CE Objective}
\label{sec:appendix_default_multitf_ce}

The default MultiTF objective is masked-token cross-entropy on masked positions in the noisy part of $\mathbf{X}_{\lambda}$.
Let
\begin{equation}
\mathcal{M}_{\lambda}
=
\{i: \mathbf{x}^{\lambda}_{\mathbf{t}}[i]=\texttt{[M]}\}
\end{equation}
denote the masked positions.
The objective is
\begin{equation}
\begin{aligned}
\mathcal{L}_{\mathrm{MultiTF}}(\theta)
=
-\mathbb{E}_{\lambda,\mathbf{t},\mathbf{x}_0}
\bigg[
\frac{1}{|\mathcal{M}_{\lambda}|}
\sum_{i\in\mathcal{M}_{\lambda}}
\log p_\theta
\bigl(x_0^i\mid\mathbf{X}_{\lambda},\mathbf{A}_{\lambda}\bigr)
\bigg].
\end{aligned}
\label{eq:appendix_multitf_ce}
\end{equation}
This objective is used for BD-LMs whose original training recipe is standard masked-token CE.

\subsubsection{DMax-enhanced Models: OPUT Self-denoising}
\label{sec:appendix_dmax_oput}

For DMax-enhanced models, we keep the same MultiTF input sequences and add the DMax OPUT self-denoising branch.
For each MultiTF input sequence, OPUT forms two branches.
The standard branch computes the training loss on the original noisy input sequence.
The self-denoising branch first runs a no-gradient forward pass, replaces masked positions in the noisy part with the model's argmax predictions, and then computes the loss on this partially self-denoised input.
Gradients flow only through the second forward pass of the self-denoising branch.
This exposes the model to partially self-generated states while keeping the MultiTF layout and attention-mask construction unchanged.
The procedure is summarized in Algorithm~\ref{alg:oput_rollout}.

\begin{algorithm}[t]
\footnotesize
\caption{DMax OPUT Self-Denoising Branch}
\label{alg:oput_rollout}
\begin{algorithmic}[1]
\Require Model $\theta$; input sequence $\mathbf{X}_{\lambda}=[\mathbf{x}^{\lambda}_{\mathbf{t}};\mathbf{x}_0]$; noisy length $N$; mask token id $m$.
\State Run a no-gradient forward pass on the noisy part: $\mathbf{L}\gets\theta(\mathbf{X}_{\lambda})_{:N}$.
\State Compute argmax predictions $\hat{\mathbf{x}}\gets\arg\max \mathbf{L}$.
\State Replace masked positions in $\mathbf{x}^{\lambda}_{\mathbf{t}}$ with $\hat{\mathbf{x}}$.
\State \Return the partially self-denoised input sequence.
\end{algorithmic}
\end{algorithm}

\subsubsection{SDAR Models: Block-wise Noise-weighted CE}
\label{sec:appendix_sdar_objective}

For SDAR models, we also reuse the same MultiTF input sequences and \textit{Group-Aware Dual-Stream Mask}s.
The difference lies in the loss normalization.
Instead of computing one global masked-token CE over all masked positions, SDAR applies a block-wise noise-weighted CE, where the loss of each block is normalized by the mask ratio applied to that block.

Let $\mathcal{B}_k$ denote token positions of block $k$ in the noisy part, and let
\begin{equation}
\mathcal{M}_{\lambda,k}
=
\mathcal{M}_{\lambda}\cap \mathcal{B}_k
\end{equation}
be the masked positions in block $k$ under layout $\lambda$.
Let $t_{\lambda,k}$ denote the mask ratio assigned to block $k$.
The SDAR-style MultiTF objective is
\begin{equation}
\begin{aligned}
\mathcal{L}_{\mathrm{MultiTF}}^{\mathrm{SDAR}}(\theta)
=
-\mathbb{E}_{\lambda,\mathbf{t},\mathbf{x}_0}
\bigg[
\frac{1}{K}
\sum_{k=1}^{K}
\frac{1}{\max(t_{\lambda,k},\epsilon)}
\sum_{i\in\mathcal{M}_{\lambda,k}}
\log p_\theta
\bigl(x_0^i\mid\mathbf{X}_{\lambda},\mathbf{A}_{\lambda}\bigr)
\bigg],
\end{aligned}
\label{eq:appendix_sdar_multitf_loss}
\end{equation}
where $\epsilon$ is a small constant used for numerical stability.
This block-wise normalization extends the diffusion loss to the full block sequence while preserving the per-block noise weighting used by SDAR.
It differs from Equation~\ref{eq:appendix_multitf_ce}, which normalizes the loss globally over all masked positions in the noisy part.

\subsection{Sorted-uniform Scheduler Baseline}
\label{sec:appendix_sorted_uniform_scheduler}

The sorted-uniform \textit{noise-scheduler} is a baseline for constructing monotonic block-level noise within each \textit{noise-group}.
For a \textit{noise-group} $H_m=(j_1,\ldots,j_{n_m})$, it independently samples $n_m$ mask ratios from a uniform distribution and then sorts them in ascending order before assigning them to the blocks in the \textit{noise-group}, as summarized in Algorithm~\ref{alg:sorted_uniform_scheduler}:
\[
u_1,\ldots,u_{n_m}
\overset{\mathrm{i.i.d.}}{\sim}
\mathcal{U}(t_{\mathrm{low}},t_{\mathrm{high}}),
\qquad
u_{(1)}\le \cdots \le u_{(n_m)},
\]
\[
t_{j_i}=u_{(i)},
\qquad
i=1,\ldots,n_m.
\]
This produces a monotonic noise pattern similar in spirit to D2F.
However, unlike the chain-uniform \textit{noise-scheduler} in Appendix~\ref{sec:appendix_chain_uniform_scheduler}, the gaps between adjacent slots are only induced by order statistics of uniformly sampled values and are not explicitly encouraged to be large.

\begin{algorithm}[t]
\footnotesize
\caption{Sorted-uniform Block-level Noise-Scheduler}
\label{alg:sorted_uniform_scheduler}
\begin{algorithmic}[1]
\Require \textit{Noise-group} $H_m=(j_1,\ldots,j_{n_m})$; noise bounds $t_{\mathrm{low}},t_{\mathrm{high}}$.
\For{$i\gets 1$ \textbf{to} $n_m$}
    \State Sample $u_i\sim\mathcal{U}(t_{\mathrm{low}},t_{\mathrm{high}})$.
\EndFor
\State Sort sampled ratios: $u_{(1)}\le\cdots\le u_{(n_m)}$.
\For{$i\gets 1$ \textbf{to} $n_m$}
    \State Assign $t_{j_i}\gets u_{(i)}$.
\EndFor
\State \Return block-level mask ratios $\{t_{j_i}\}_{i=1}^{n_m}$.
\end{algorithmic}
\end{algorithm}

%% file: sections/10_appendix_system_design_detail.tex
\section{MultiBD Inference Implementation Details}
\label{sec:appendix_system}

This appendix expands the optimized MultiBD inference algorithm introduced in Section~\ref{sec:decoding}.
The main design goal is to execute the MultiBD \textit{running-set} in Equation~\ref{eq:multibd_state} with a static physical input shape, while preserving prefix KV-cache reuse.


\begin{figure*}[t]
  \centering

  \begin{subfigure}[t]{0.48\textwidth}
    \centering
    \includegraphics[width=\linewidth]{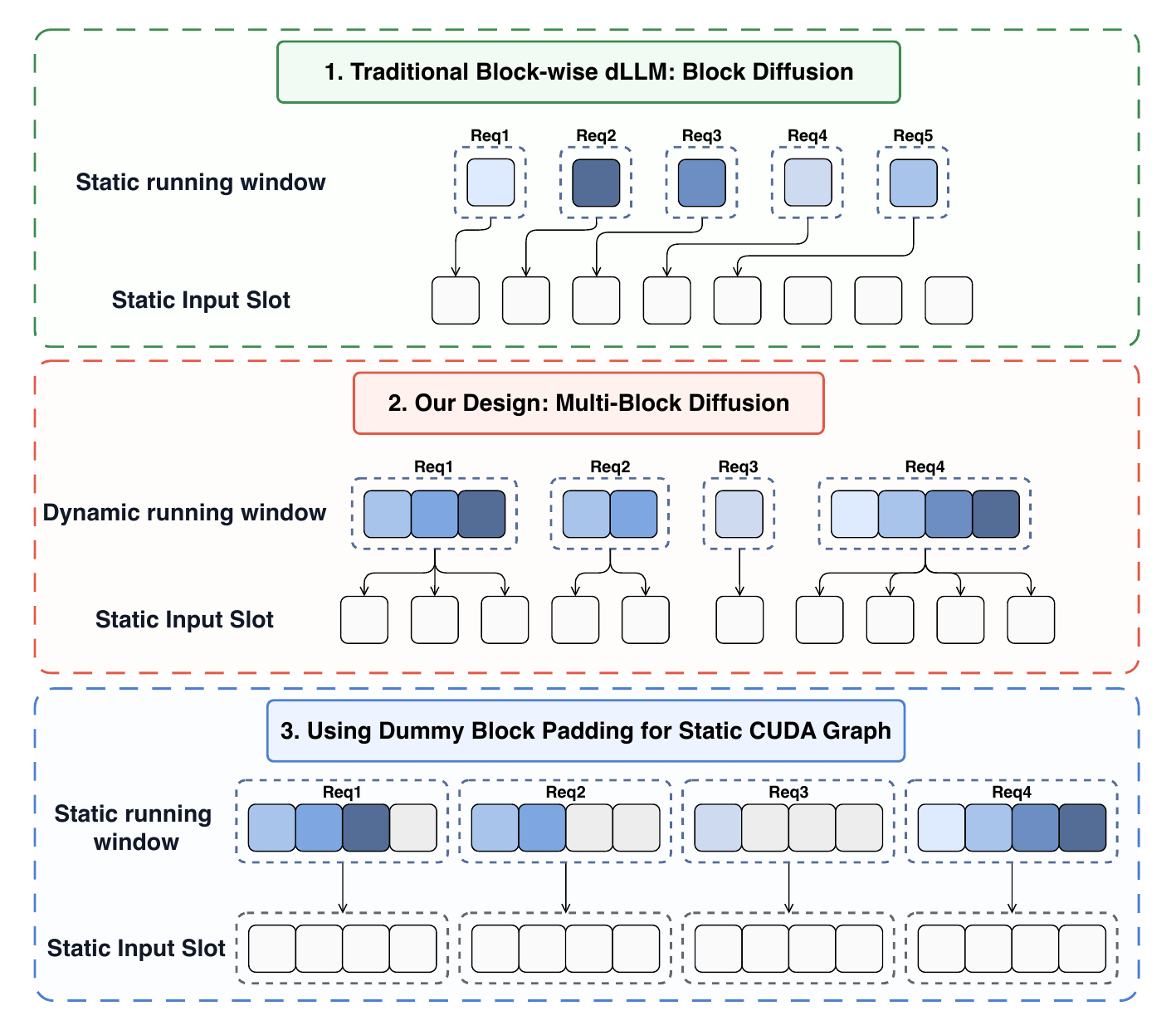}
    \caption{
    CUDA Graph compatibility across decoding designs.
    \textbf{(1)} SingleBD uses a fixed single active block but exposes no inter-block parallelism.
    \textbf{(2)} Naive MultiBD appends future blocks dynamically, making the \textit{running-set} length change over time.
    \textbf{(3)} Optimized MultiBD maps the logical \textit{running-set} into a fixed-size \textit{Block Buffer} with dummy slots, keeping tensor shapes static for CUDA Graph capture and replay.
    }
    \label{fig:cuda_graph}
  \end{subfigure}
  \hfill
  \begin{subfigure}[t]{0.48\textwidth}
    \centering
    \includegraphics[width=\linewidth]{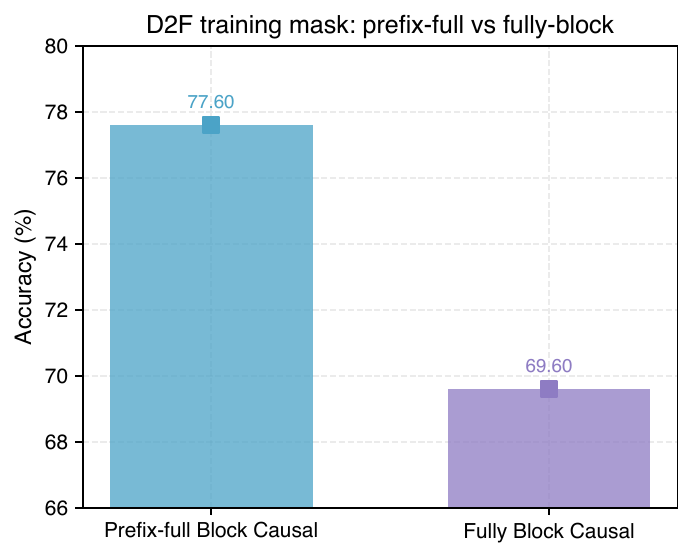}
    \caption{
    Making D2F fully block-causal hurts accuracy.
    Prefix-full attention gives D2F stronger noisy-prefix visibility but is not naturally compatible with prefix KV caching.
    Directly replacing it with a fully block-causal mask improves cache compatibility but drops accuracy from 77.60\% to 69.60\%.
    }
    \label{fig:d2f_prefix_cache_ablation}
  \end{subfigure}

  \caption{
  Static-shape execution and prefix-cache compatibility analyses.
  Left: optimized MultiBD keeps tensor shapes static through a fixed-size \textit{Block Buffer}, enabling CUDA Graph capture and replay.
  Right: making D2F fully block-causal improves cache compatibility but substantially hurts accuracy.
  }
  \label{fig:cuda_graph_and_d2f_ablation}
\end{figure*}

\subsection{A dynamic \textit{running-set} prevents static-shape execution.}
\label{sec:appendix_naive_multibd}

A direct implementation of MultiBD maintains a dynamic \textit{running-set} in addition to the committed prefix cache.
When the latest active block reaches an add-block threshold, the decoder appends a fully masked future block to the \textit{running-set}.
When the front active block is completed, the decoder writes it into the KV cache and removes it from the \textit{running-set}.
This dynamic procedure exposes inter-block parallelism, but the number of active tokens changes across decoding steps and across requests.
As shown in Figure~\ref{fig:cuda_graph}(2), such shape variation is unfriendly to CUDA Graph capture and replay.

\begin{algorithm}[t]
\footnotesize
\caption{Naive MultiBD with a Dynamic Running-Set}
\label{alg:naive_multibd}
\begin{algorithmic}[1]
\Require Model $\theta$; block size $B$; thresholds $\tau_{\mathrm{add}}$, $\tau_{\mathrm{semi}}$, $\tau_{\mathrm{M2T}}$.

\Statex \textcolor{commentcolor}{// \textbf{Initialize dynamic MultiBD state}}
\State Initialize prefix KV cache $\mathcal{K}\gets\emptyset$ and dynamic \textit{running-set} $\mathcal{Y}\gets\emptyset$.
\State Append one fully masked \textsc{active} block to $\mathcal{Y}$.

\While{generation is not complete}
    \Statex \hspace{\algorithmicindent}\textcolor{commentcolor}{// \textbf{Grow the \textit{running-set} dynamically}}
    \If{the latest active block has progress $>\tau_{\mathrm{add}}$ and EOS has not appeared}
        \State Append a fully masked future block to $\mathcal{Y}$.
    \EndIf

    \Statex \hspace{\algorithmicindent}\textcolor{commentcolor}{// \textbf{Decode all blocks in the current \textit{running-set}}}
    \State Run $\theta$ on $\mathcal{Y}$ with prefix cache $\mathcal{K}$.
    \For{each active block $b\in\mathcal{Y}$}
        \State Accept masked positions with confidence $>\tau_{\mathrm{M2T}}$.
        \If{the previous active block is semi-complete and no token is accepted}
            \State Accept the highest-confidence masked position.
        \EndIf
        \If{$b$ is fully decoded}
            \State Mark $b$ as \textsc{to-cache}.
        \EndIf
    \EndFor

    \Statex \hspace{\algorithmicindent}\textcolor{commentcolor}{// \textbf{Commit completed prefix blocks}}
    \While{the front block of $\mathcal{Y}$ is \textsc{to-cache}}
        \State Write the front block into $\mathcal{K}$ and remove it from $\mathcal{Y}$.
    \EndWhile
\EndWhile
\State \Return generated tokens.
\end{algorithmic}
\end{algorithm}

\subsection{A fixed \textit{Block Buffer} implements MultiBD states.}
\label{sec:appendix_fixed_buffer}

Optimized MultiBD replaces dynamic appending with a fixed-size \textit{Block Buffer}.
The \textit{Block Buffer} contains $N_{\mathrm{buf}}$ physical block slots.
At each decoding step, active slots represent the logical \textit{running-set} $\mathcal{R}_s$, while dummy slots reserve capacity for future blocks.
Adding a future block therefore activates an existing dummy slot instead of extending the physical input sequence.
When the front active block is completed, it is committed to the KV cache, removed from the \textit{running-set}, and the \textit{Block Buffer} slides forward by replacing the consumed slot with a new dummy slot at the tail.
This realizes MultiBD while keeping the number of processed buffer tokens fixed at $N_{\mathrm{buf}}\cdot B$.

\begin{algorithm}[t]
\footnotesize
\caption{Optimized MultiBD with a Fixed \textit{Block Buffer}}
\label{alg:optimized_multibd}
\begin{algorithmic}[1]
\Require Model $\theta$; block size $B$; buffer size $N_{\mathrm{buf}}$; thresholds $\tau_{\mathrm{add}}$, $\tau_{\mathrm{semi}}$, $\tau_{\mathrm{stable}}$, $\tau_{\mathrm{M2T}}$, and optional $\tau_{\mathrm{T2T}}$.

\Statex \textcolor{commentcolor}{// \textbf{Initialize fixed \textit{Block Buffer}}}
\State Initialize prefix KV cache $\mathcal{K}$ and a fixed \textit{Block Buffer} $\mathcal{W}$ with $N_{\mathrm{buf}}$ slots.
\State Set $\mathcal{W}[0]$ to a fully masked \textsc{active} block and all remaining slots to \textsc{dummy}.

\While{generation is not complete}
    \Statex \hspace{\algorithmicindent}\textcolor{commentcolor}{// \textbf{Activate future blocks without changing shape}}
    \State Let $\mathcal{R}$ be the non-\textsc{dummy} resident blocks in $\mathcal{W}$.
    \State Let $b_{\mathrm{last}}$ be the last \textsc{active} block in $\mathcal{R}$.
    \If{$b_{\mathrm{last}}$ satisfies progress $>\tau_{\mathrm{add}}$ and stability $>\tau_{\mathrm{stable}}$}
        \State Activate the first trailing \textsc{dummy} slot if one exists.
    \EndIf

    \Statex \hspace{\algorithmicindent}\textcolor{commentcolor}{// \textbf{Decode the static \textit{Block Buffer}}}
    \State Run $\theta$ on the static $N_{\mathrm{buf}}\cdot B$ \textit{Block Buffer} tokens with prefix cache $\mathcal{K}$.
    \For{each \textsc{active} block $b\in\mathcal{W}$}
        \State Accept masked positions with confidence $>\tau_{\mathrm{M2T}}$.
        \If{no masked position is accepted and the preceding active block is semi-complete}
            \State Accept the highest-confidence masked position in $b$.
        \EndIf
        \If{T2T revision is enabled}
            \State Revise eligible filled but uncommitted positions with confidence $>\tau_{\mathrm{T2T}}$.
        \EndIf
        \If{$b$ is complete and all preceding resident blocks are cached or ready-to-cache}
            \State Mark $b$ as \textsc{to-cache}.
        \EndIf
    \EndFor

    \Statex \hspace{\algorithmicindent}\textcolor{commentcolor}{// \textbf{Commit prefix blocks and slide the buffer}}
    \While{the front slot of $\mathcal{W}$ is \textsc{to-cache}}
        \State Write the front block into $\mathcal{K}$; its state becomes \textsc{in-cache}.
        \State Pop the front slot and append a new \textsc{dummy} slot at the tail.
    \EndWhile
\EndWhile
\State \Return generated tokens.
\end{algorithmic}
\end{algorithm}

\begin{figure*}[t]
  \centering
  \includegraphics[width=\textwidth]{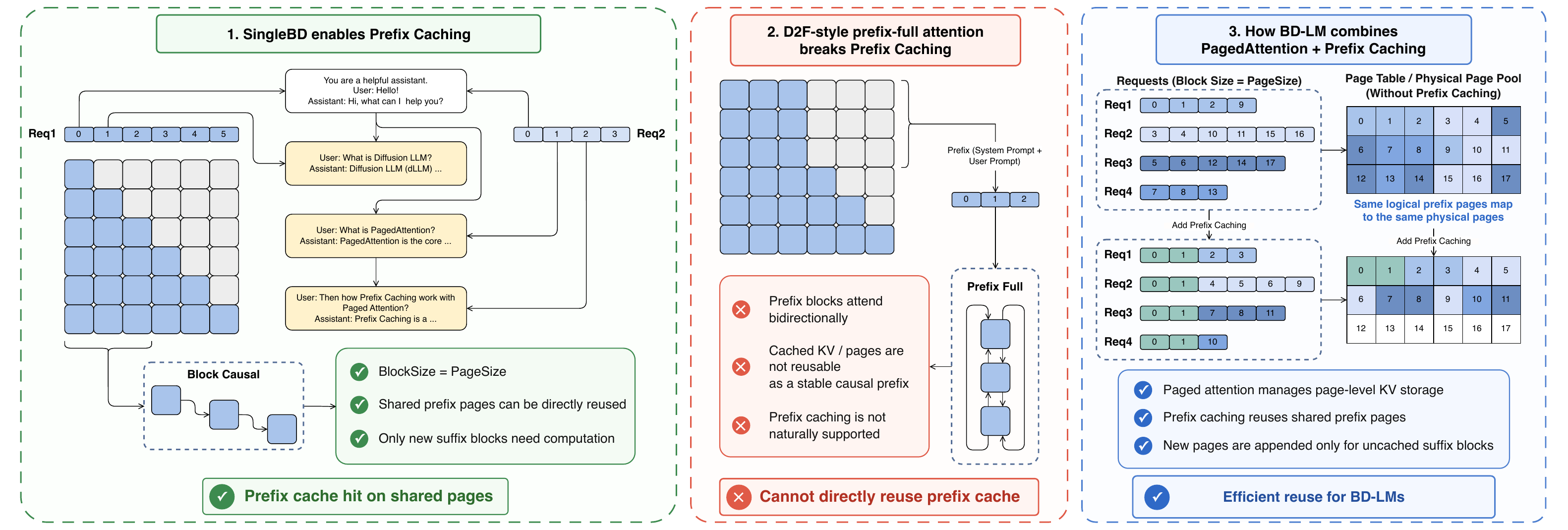}
  \caption{
    Prefix caching in block-causal BD-LMs.
    \textbf{(1)} SingleBD keeps completed blocks as an immutable clean prefix, enabling direct KV-cache reuse.
    \textbf{(2)} D2F-style prefix-full attention breaks this cache semantics because noisy prefix blocks are not reusable as stable causal prefix pages.
    \textbf{(3)} \textit{Block Buffer} MultiBD separates cached prefix blocks from active \textit{Block Buffer} slots, enabling prefix KV reuse while refining multiple active blocks.
    }
  \label{fig:prefix_caching}
\end{figure*}

\subsection{Block states advance the fixed \textit{Block Buffer}.}
\label{sec:appendix_block_states}

Each physical slot in the \textit{Block Buffer} follows the transition
\[
\textsc{dummy} \rightarrow \textsc{active} \rightarrow \textsc{to-cache} \rightarrow \textsc{in-cache}.
\]
A \textsc{dummy} slot is an idle placeholder that preserves the static buffer shape.
An \textsc{active} slot participates in the current MultiBD forward pass.
A \textsc{to-cache} block has completed decoding and is ready to be committed.
An \textsc{in-cache} block has been written into the prefix KV cache and no longer belongs to the active part of the \textit{running-set}.
These state transitions implement the logical evolution of $\mathcal{R}_s$ without changing the physical input shape.


\subsection{Thresholds control activation and token updates.}
\label{sec:appendix_thresholds}

MultiBD uses separate thresholds for block activation, fallback progress, and token updates.
The add-block threshold $\tau_{\mathrm{add}}$ controls when a future block can enter the fixed \textit{Block Buffer}.
The stability threshold $\tau_{\mathrm{stable}}$ prevents premature activation when the current latest active block is still unstable.
The semi-completion threshold $\tau_{\mathrm{semi}}$ allows later active blocks to use the top-1 context of a preceding block once it has made sufficient progress, even before it is fully cached.
The M2T threshold $\tau_{\mathrm{M2T}}$ controls mask-to-token acceptance, and the optional T2T threshold $\tau_{\mathrm{T2T}}$ controls token-to-token revision for models that support T2T updates.

This separation is important because M2T and T2T updates have different reliability profiles.
M2T introduces new content into an active block, while T2T overwrites tentative content before commitment.
Using separate thresholds stabilizes concurrent block refinement and reduces error propagation across the \textit{running-set}.

\subsection{Prefix Caching and Fully Block-Causal D2F}
\label{sec:appendix_prefix_cache}

\paragraph{Native D2F is not directly prefix-cache compatible.}
Prefix caching is a key advantage of BD-LMs.
In SingleBD, completed blocks form an immutable clean prefix, so their KV states can be stored and directly reused in later steps.
As shown in Figure~\ref{fig:prefix_caching}(1), only the current noisy block requires repeated computation.
By contrast, native D2F uses prefix-full attention.
Future noisy blocks condition on a prefix-full context, where prefix states are not organized as immutable block-causal prefix pages in the standard BD-LM cache.
As illustrated in Figure~\ref{fig:prefix_caching}(2), their KV states cannot be reused in the same way as SingleBD prefix blocks.

\paragraph{Fully block-causal D2F variant.}
To isolate the prefix-caching issue, we construct a fully block-causal D2F variant.
Let the full clean sequence be partitioned into BD-LM blocks:
\[
\mathbf{x}_0=[\mathbf{b}_1,\ldots,\mathbf{b}_K],
\qquad
\mathbf{b}_k\in\mathcal{V}^{B}.
\]
Suppose native D2F uses a token-level clean prefix
\[
\mathbf{x}_0^{\mathrm{pre}}=(x_0^1,\ldots,x_0^P),
\]
where $P$ can be arbitrary and need not be divisible by $B$.
Let
\[
a=\left\lfloor \frac{P}{B}\right\rfloor+1,
\qquad
r=P-(a-1)B
\]
denote the first block that contains suffix tokens and the number of prefix tokens inside this boundary block, respectively.
Then $\mathbf{b}_1,\ldots,\mathbf{b}_{a-1}$ are complete clean prefix blocks, while $\mathbf{b}_a$ may contain both prefix tokens and suffix tokens.

We use $\mathbf{b}_a$ as the first noisy block of the D2F-style suffix, rather than inserting padding tokens.
For the boundary block, only its suffix positions are noised and included in the loss:
\[
\mathcal{I}_a=\{r+1,\ldots,B\}.
\]
For later blocks $j>a$, all positions belong to the suffix:
\[
\mathcal{I}_j=\{1,\ldots,B\}.
\]
We then apply a monotonic D2F-style \textit{noise-scheduler} to the valid suffix positions of blocks $a,\ldots,K$:
\[
0\le t_a<t_{a+1}<\cdots<t_K\le 1.
\]
Let $\bar{\mathbf{b}}_{j,t_j}$ denote the partially corrupted block, where positions in $\mathcal{I}_j$ are corrupted by $q_{t_j}(\cdot\mid \mathbf{b}_j)$ and positions outside $\mathcal{I}_j$ are kept clean.
For the boundary block, this means that the prefix part of $\mathbf{b}_a$ remains clean, while the suffix part is noised.

The fully block-causal D2F variant factorizes the suffix as
\begin{equation}
\begin{aligned}
p_\theta
\bigl(
\mathbf{x}_0^{\mathrm{suf}}
\mid
\mathbf{x}_0^{(<a)},
\bar{\mathbf{b}}_{a,t_a},\ldots,\bar{\mathbf{b}}_{K,t_K}
\bigr)
=
\prod_{j=a}^{K}
p_\theta
\bigl(
\mathbf{b}_j^{\mathcal{I}_j}
\mid
\mathbf{x}_0^{(<a)},
\bar{\mathbf{b}}_{a,t_a},\ldots,\bar{\mathbf{b}}_{j,t_j}
\bigr),
\end{aligned}
\label{eq:fully_block_causal_d2f}
\end{equation}
where $\mathbf{x}_0^{(<a)}=[\mathbf{b}_1,\ldots,\mathbf{b}_{a-1}]$ is the block-causal clean prefix and $\mathbf{b}_j^{\mathcal{I}_j}$ denotes the suffix positions of block $j$.
The loss is computed only on masked positions within $\mathcal{I}_j$.

Compared with native D2F, this variant changes the training-state construction by replacing prefix-full attention with a fully block-causal attention.
Equivalently, it completes the arbitrary token-level prefix to the next block boundary using real continuation tokens from the training sequence, and treats the boundary block as the first block in the noisy suffix.
This construction is the training-side counterpart of the extreme MBD-LM state discussed in Section~\ref{sec:mbd_lms}, where the \textit{running-set} covers all suffix blocks and follows a monotonic D2F-style \textit{noise-scheduler}.

\paragraph{Fully block-causal D2F is not a sufficient fix.}
Although the fully block-causal variant improves cache compatibility, it substantially hurts accuracy.
As shown in Figure~\ref{fig:d2f_prefix_cache_ablation}, changing D2F from prefix-full attention to fully block-causal attention drops accuracy from 77.60\% to 69.60\%.
This suggests that D2F relies on stronger prefix-full visibility, and cache compatibility cannot be obtained by simply restricting the attention mask.
This result further motivates MultiTF, which keeps the block-causal cached-prefix interface while training on bounded \textit{noise-groups} that better match MultiBD inference.

\paragraph{\textit{Block Buffer} MultiBD preserves cache semantics.}
Our \textit{Block Buffer} MultiBD design preserves prefix caching by construction.
Committed blocks become immutable \textsc{in-cache} prefix context and are represented only through cached KV states.
Active blocks remain inside the \textit{Block Buffer} and are recomputed during iterative refinement, while future \textsc{dummy} slots remain invisible until activated.
As shown in Figure~\ref{fig:prefix_caching}(3), this separates cached prefix blocks from active \textit{Block Buffer} slots, enabling prefix KV reuse while still refining multiple active blocks in parallel.

%% file: sections/11_appendix_experiments.tex
\section{Experimental Details}
\label{sec:appendix_exp_details}

This appendix reports the inference and MultiTF post-training hyperparameters used in our experiments.
``---'' indicates that the corresponding hyperparameter is not applicable.
SingleBD (Native) denotes the original single-block inference of each BD-LM; MultiBD (training-free) denotes MultiBD inference without post-training; MBD-* denotes the corresponding MultiTF-post-trained model.

\begin{table*}[t]
\centering
\small
\caption{
Inference hyperparameters for all evaluated configurations.
$\tau_{\mathrm{add}}$ controls when a future block is activated;
$\tau_{\mathrm{semi}}$ controls semi-completion or fallback progress;
$\tau_{\mathrm{stable}}$ controls activation stability;
$\tau_{\mathrm{M2T}}$ and $\tau_{\mathrm{T2T}}$ are confidence thresholds for mask-to-token filling and token-to-token revision.
}
\label{tab:appendix_infer_hparams}
\resizebox{\textwidth}{!}{%
\begin{tabular}{llcccccccccc}
\toprule
\textbf{Configuration}
& \textbf{Task}
& \textbf{Buffer}
& \textbf{Block}
& \textbf{Max Len}
& \textbf{Max New}
& \textbf{Max NFE}
& $\boldsymbol{\tau_{\mathrm{add}}}$
& $\boldsymbol{\tau_{\mathrm{semi}}}$
& $\boldsymbol{\tau_{\mathrm{stable}}}$
& $\boldsymbol{\tau_{\mathrm{M2T}}}$
& $\boldsymbol{\tau_{\mathrm{T2T}}}$ \\
\midrule

\multicolumn{12}{l}{\textbf{\textit{LLaDA2-Mini-DMax}}} \\
SingleBD (Native)          & Math & 1 & 32 & 4096 & 4096 & 1024 & ---  & ---  & ---  & 0.50 & --- \\
SingleBD (Native)          & Code & 1 & 32 & 4096 & 4096 & 1024 & ---  & ---  & ---  & 0.65 & --- \\
MultiBD (training-free)    & Math & 2 & 32 & 4096 & 4096 & 1024 & 0.10 & 0.90 & 0.50 & 0.50 & --- \\
MultiBD (training-free)    & Code & 2 & 32 & 4096 & 4096 & 1024 & 0.90 & 0.90 & 0.50 & 0.65 & --- \\
MBD-LLaDA2-Mini-DMax       & Math & 2 & 32 & 4096 & 4096 & 1024 & 0.10 & 0.90 & 0.50 & 0.50 & --- \\
MBD-LLaDA2-Mini-DMax       & Code & 2 & 32 & 4096 & 4096 & 1024 & 0.90 & 0.90 & 0.50 & 0.65 & --- \\
\midrule

\multicolumn{12}{l}{\textbf{\textit{LLaDA2-Mini}}} \\
SingleBD (Native)          & Math & 1 & 32 & 4096 & 4096 & 1024 & ---  & ---  & --- & 0.95 & --- \\
SingleBD (Native)          & Code & 1 & 32 & 4096 & 4096 & 1024 & ---  & ---  & --- & 0.95 & --- \\
MultiBD (training-free)    & Math & 2 & 32 & 4096 & 4096 & 1024 & 0.10 & 0.90 & --- & 0.95 & --- \\
MultiBD (training-free)    & Code & 2 & 32 & 4096 & 4096 & 1024 & 0.90 & 0.90 & --- & 0.95 & --- \\
MBD-LLaDA2-Mini            & Math & 2 & 32 & 4096 & 4096 & 1024 & 0.10 & 0.90 & --- & 0.95 & --- \\
MBD-LLaDA2-Mini            & Code & 2 & 32 & 4096 & 4096 & 1024 & 0.90 & 0.90 & --- & 0.95 & --- \\
\midrule

\multicolumn{12}{l}{\textbf{\textit{SDAR-8B-Chat-b32}}} \\
SingleBD (Native)          & Math & 1 & 32 & 4096 & 4096 & 1024 & ---  & ---  & --- & 0.95 & --- \\
SingleBD (Native)          & Code & 1 & 32 & 4096 & 4096 & 1024 & ---  & ---  & --- & 0.95 & --- \\
MultiBD (training-free)    & Math & 4 & 32 & 4096 & 4096 & 1024 & 0.10 & 0.90 & --- & 0.95 & --- \\
MultiBD (training-free)    & Code & 4 & 32 & 4096 & 4096 & 1024 & 0.90 & 0.90 & --- & 0.95 & --- \\
MBD-SDAR-8B-Chat-b32       & Math & 4 & 32 & 4096 & 4096 & 1024 & 0.10 & 0.90 & --- & 0.95 & --- \\
MBD-SDAR-8B-Chat-b32       & Code & 4 & 32 & 4096 & 4096 & 1024 & 0.90 & 0.90 & --- & 0.95 & --- \\
\midrule

\multicolumn{12}{l}{\textbf{\textit{SDAR-8B-Chat-b4}}} \\
SingleBD (Native)          & Math & 1 & 4 & 4096 & 4096 & 1024 & ---  & ---  & --- & 0.95 & --- \\
SingleBD (Native)          & Code & 1 & 4 & 4096 & 4096 & 1024 & ---  & ---  & --- & 0.95 & --- \\
MultiBD (training-free)    & Math & 4 & 4 & 4096 & 4096 & 1024 & 0.10 & 0.25 & --- & 0.95 & --- \\
MultiBD (training-free)    & Code & 4 & 4 & 4096 & 4096 & 1024 & 0.75 & 0.75 & --- & 0.95 & --- \\
MBD-SDAR-8B-Chat-b4        & Math & 4 & 4 & 4096 & 4096 & 1024 & 0.10 & 0.25 & --- & 0.95 & --- \\
MBD-SDAR-8B-Chat-b4        & Code & 4 & 4 & 4096 & 4096 & 1024 & 0.75 & 0.75 & --- & 0.95 & --- \\
\midrule

\multicolumn{12}{l}{\textbf{\textit{LLaDA2-Mini-CAP}}} \\
SingleBD (Native)          & Math & 1 & 32 & 4096 & 4096 & 1024 & ---  & ---  & --- & 0.95 & --- \\
SingleBD (Native)          & Code & 1 & 32 & 4096 & 4096 & 1024 & ---  & ---  & --- & 0.95 & --- \\
MultiBD (training-free)    & Math & 2 & 32 & 4096 & 4096 & 1024 & 0.10 & 0.90 & --- & 0.95 & --- \\
MultiBD (training-free)    & Code & 2 & 32 & 4096 & 4096 & 1024 & 0.90 & 0.90 & --- & 0.95 & --- \\
\midrule

\multicolumn{12}{l}{\textbf{\textit{LLaDA2.1-Mini}}} \\
SingleBD (Native)          & Math & 1 & 32 & 4096 & 4096 & 1024 & ---  & ---  & --- & 0.70 & 0.50 \\
SingleBD (Native)          & Code & 1 & 32 & 4096 & 4096 & 1024 & ---  & ---  & --- & 0.70 & 0.50 \\
MultiBD (training-free)    & Math & 2 & 32 & 4096 & 4096 & 1024 & 0.10 & 0.90 & --- & 0.70 & 0.50 \\
MultiBD (training-free)    & Code & 2 & 32 & 4096 & 4096 & 1024 & 0.90 & 0.90 & --- & 0.70 & 0.50 \\

\bottomrule
\end{tabular}%
}
\end{table*}

\begin{table*}[t]
\centering
\small
\caption{ 
MultiTF post-training hyperparameters.
$t_{\mathrm{low}}$ and $t_{\mathrm{high}}$ denote the mask-ratio range;
$\rho$ is the margin ratio used to determine the effective upper bound $t_{\mathrm{eff}}$;
$N_{\mathrm{rand}}$ is the number of random \textit{group-layouts} per sample.
The random-scheduler ablation uses a separate power-law bias $\gamma_{\mathrm{rand}}$, which is independent of $\rho$ and is not used in the chain-uniform scheduler.
}
\label{tab:appendix_train_hparams}
\resizebox{\textwidth}{!}{%
\begin{tabular}{llcccccccccc}
\toprule
\textbf{Target Model}
& \textbf{Task}
& \textbf{Objective}
& \textbf{Data}
& \textbf{Seq Len}
& \textbf{Block}
& \textbf{Max Group}
& $\boldsymbol{t_{\mathrm{low}}}$
& $\boldsymbol{t_{\mathrm{high}}}$
& $\boldsymbol{\rho}$
& $\boldsymbol{N_{\mathrm{rand}}}$
& \textbf{Steps} \\
\midrule

MBD-LLaDA2-Mini-DMax & Math 
& MultiTF + DMax OPUT & 60k & 2048 & 32 & 2 
& 0.001 & 1.00 & $\rho_{\mathrm{cfg}}$ & 0 & 15000 \\
MBD-LLaDA2-Mini-DMax & Code 
& MultiTF + DMax OPUT & 60k & 2048 & 32 & 2 
& 0.001 & 1.00 & $\rho_{\mathrm{cfg}}$ & 2 & 4000 \\
\midrule

MBD-LLaDA2-Mini & Math 
& MultiTF CE & 60k & 2048 & 32 & 2 
& 0.001 & 1.00 & $\rho_{\mathrm{cfg}}$ & 0 & 15000 \\
MBD-LLaDA2-Mini & Code 
& MultiTF CE & 60k & 2048 & 32 & 2 
& 0.001 & 1.00 & $\rho_{\mathrm{cfg}}$ & 0 & 6500 \\
\midrule

MBD-SDAR-8B-Chat-b32 & Math 
& MultiTF CE & 20k & 2048 & 32 & 4 
& 0.001 & 1.00 & $\rho_{\mathrm{cfg}}$ & 3 & 3125 \\
MBD-SDAR-8B-Chat-b32 & Code 
& MultiTF CE & 10k & 2048 & 32 & 4 
& 0.001 & 1.00 & $\rho_{\mathrm{cfg}}$ & 3 & 1670 \\
\midrule

MBD-SDAR-8B-Chat-b4 & Math 
& MultiTF CE & 20k & 2048 & 4 & 4 
& 0.001 & 1.00 & $\rho_{\mathrm{cfg}}$ & 2 & 1250 \\
MBD-SDAR-8B-Chat-b4 & Code 
& MultiTF CE & 10k & 2048 & 4 & 4 
& 0.001 & 1.00 & $\rho_{\mathrm{cfg}}$ & 2 & 200 \\

\bottomrule
\end{tabular}%
}
\end{table*}

%% file: main.bbl
\begin{thebibliography}{21}
\providecommand{\natexlab}[1]{#1}
\providecommand{\url}[1]{\texttt{#1}}
\expandafter\ifx\csname urlstyle\endcsname\relax
  \providecommand{\doi}[1]{doi: #1}\else
  \providecommand{\doi}{doi: \begingroup \urlstyle{rm}\Url}\fi

\bibitem[Arriola et~al.(2025)Arriola, Gokaslan, Chiu, Yang, Qi, Han, Sahoo, and Kuleshov]{arriola2025block}
Marianne Arriola, Aaron Gokaslan, Justin~T. Chiu, Zhihan Yang, Zhixuan Qi, Jiaqi Han, Subham~Sekhar Sahoo, and Volodymyr Kuleshov.
\newblock Block diffusion: Interpolating between autoregressive and diffusion language models.
\newblock In \emph{International Conference on Learning Representations (ICLR)}, 2025.
\newblock URL \url{https://arxiv.org/abs/2503.09573}.
\newblock Oral Presentation.

\bibitem[Bie et~al.(2025)Bie, Huang, Li, et~al.]{bie2025llada2}
Tiwei Bie, Zenan Huang, Chongxuan Li, et~al.
\newblock Llada2.0: Scaling up diffusion language models to 100b.
\newblock \emph{arXiv preprint arXiv:2512.15745}, 2025.
\newblock URL \url{https://arxiv.org/abs/2512.15745}.

\bibitem[Bie et~al.(2026)]{bie2026llada21}
Tiwei Bie et~al.
\newblock Llada2.1: Speeding up text diffusion via token editing.
\newblock \emph{arXiv preprint arXiv:2602.08676}, 2026.
\newblock URL \url{https://arxiv.org/abs/2602.08676}.

\bibitem[Boizard et~al.(2025)Boizard, Gisserot-Boukhlef, El-Haddad, Hudelot, and Colombo]{boizard2025reasoning}
Nicolas Boizard, Hippolyte Gisserot-Boukhlef, Kevin El-Haddad, C{\'e}line Hudelot, and Pierre Colombo.
\newblock When does reasoning matter? a controlled study of reasoning's contribution to model performance.
\newblock \emph{arXiv preprint arXiv:2509.22193}, 2025.
\newblock URL \url{https://arxiv.org/abs/2509.22193}.

\bibitem[Chen et~al.(2025)Chen, Fang, Ma, Yu, and Wang]{chen2025dparallel}
Zigeng Chen, Gongfan Fang, Xinyin Ma, Ruonan Yu, and Xinchao Wang.
\newblock dparallel: Learnable parallel decoding for dllms.
\newblock \emph{arXiv preprint arXiv:2509.26488}, 2025.
\newblock URL \url{https://arxiv.org/abs/2509.26488}.

\bibitem[Chen et~al.(2026)Chen, Fang, Ma, Yu, and Wang]{chen2026dmax}
Zigeng Chen, Gongfan Fang, Xinyin Ma, Ruonan Yu, and Xinchao Wang.
\newblock Dmax: Aggressive parallel decoding for dllms.
\newblock \emph{arXiv preprint arXiv:2604.08302}, 2026.
\newblock URL \url{https://arxiv.org/abs/2604.08302}.

\bibitem[Cheng et~al.(2025)Cheng, Bian, Liu, Zhang, Yao, Tian, Wang, Guo, Chen, Qi, and Zhou]{cheng2025sdar}
Shuang Cheng, Yihan Bian, Dawei Liu, Linfeng Zhang, Qian Yao, Zhongbo Tian, Wenhai Wang, Qipeng Guo, Kai Chen, Biqing Qi, and Bowen Zhou.
\newblock Sdar: A synergistic diffusion-autoregression paradigm for scalable sequence generation.
\newblock \emph{arXiv preprint arXiv:2510.06303}, 2025.
\newblock URL \url{https://arxiv.org/abs/2510.06303}.

\bibitem[Cobbe et~al.(2021)Cobbe, Kosaraju, Bavarian, Chen, Jun, Kaiser, Plappert, Tworek, Hilton, Nakano, Hesse, and Schulman]{gsm8k}
Karl Cobbe, Vineet Kosaraju, Mohammad Bavarian, Mark Chen, Heewoo Jun, Lukasz Kaiser, Matthias Plappert, Jerry Tworek, Jacob Hilton, Reiichiro Nakano, Christopher Hesse, and John Schulman.
\newblock Training verifiers to solve math word problems.
\newblock \emph{arXiv preprint arXiv:2110.14168}, 2021.

\bibitem[Hendrycks et~al.(2021)Hendrycks, Burns, Kadavath, Arora, Basart, Tang, Song, and Steinhardt]{math}
Dan Hendrycks, Collin Burns, Saurav Kadavath, Akul Arora, Steven Basart, Eric Tang, Dawn Song, and Jacob Steinhardt.
\newblock Measuring mathematical problem solving with the math dataset.
\newblock \emph{arXiv preprint arXiv:2103.03874}, 2021.

\bibitem[Hu et~al.(2026)Hu, Jin, Liu, Yu, and Deng]{hu2026lightningrl}
Yanzhe Hu, Yijie Jin, Pengfei Liu, Kai Yu, and Zhijie Deng.
\newblock Lightningrl: Breaking the accuracy--parallelism trade-off of block-wise dllms via reinforcement learning.
\newblock \emph{arXiv preprint arXiv:2603.13319}, 2026.
\newblock URL \url{https://arxiv.org/abs/2603.13319}.

\bibitem[{jtatman}(2025)]{jtatman2025pythoncode500k}
{jtatman}.
\newblock Python code dataset 500k.
\newblock Hugging Face dataset, 2025.
\newblock URL \url{https://huggingface.co/datasets/jtatman/python-code-dataset-500k}.

\bibitem[Liu et~al.(2023)Liu, Xia, Wang, and Zhang]{evalplus}
Jiawei Liu, Chunqiu~Steven Xia, Yuyao Wang, and Lingming Zhang.
\newblock Is your code generated by chatgpt really correct? rigorous evaluation of large language models for code generation.
\newblock \emph{Advances in Neural Information Processing Systems}, 36:\penalty0 21558--21572, 2023.

\bibitem[Lu et~al.(2026)Lu, Chen, Karashima, Wang, Fujiki, and Fan]{lu2026adablock}
Guanxi Lu, Hao~Mark Chen, Yuto Karashima, Zhican Wang, Daichi Fujiki, and Hongxiang Fan.
\newblock Adablock-dllm: Semantic-aware diffusion llm inference via adaptive block size.
\newblock \emph{arXiv preprint arXiv:2509.26432}, 2026.
\newblock URL \url{https://arxiv.org/abs/2509.26432}.

\bibitem[Ma et~al.(2025)Ma, Zheng, Shi, Zhao, Jia, Huang, Lin, Li, Yang, Peng, Zhang, and Liu]{ma2025veomni}
Qianli Ma, Yaowei Zheng, Zhelun Shi, Zhongkai Zhao, Bin Jia, Ziyue Huang, Zhiqi Lin, Youjie Li, Jiacheng Yang, Yanghua Peng, Zhi Zhang, and Xin Liu.
\newblock Veomni: Scaling any modality model training with model-centric distributed recipe zoo.
\newblock \emph{arXiv preprint arXiv:2508.02317}, 2025.
\newblock URL \url{https://arxiv.org/abs/2508.02317}.

\bibitem[Nie et~al.(2025)Nie, Zhu, You, Zhang, Ou, Hu, Zhou, Lin, Wen, and Li]{nie2025llada}
Shen Nie, Fengqi Zhu, Zebin You, Xiaolu Zhang, Jingyang Ou, Jun Hu, Jun Zhou, Yankai Lin, Ji-Rong Wen, and Chongxuan Li.
\newblock Large language diffusion models.
\newblock \emph{arXiv preprint arXiv:2502.09992}, 2025.
\newblock URL \url{https://arxiv.org/abs/2502.09992}.

\bibitem[Qian et~al.(2026)Qian, Su, Hu, Zhang, Deng, Zhao, and Zhang]{qian2026d3llm}
Yu-Yang Qian, Junda Su, Lanxiang Hu, Peiyuan Zhang, Zhijie Deng, Peng Zhao, and Hao Zhang.
\newblock d3llm: Ultra-fast diffusion llm using pseudo-trajectory distillation.
\newblock \emph{arXiv preprint arXiv:2601.07568}, 2026.
\newblock URL \url{https://arxiv.org/abs/2601.07568}.

\bibitem[Sahoo et~al.(2024)Sahoo, Arriola, Schiff, Gokaslan, Marroquin, Chiu, Rush, and Kuleshov]{sahoo2024mdlm}
Subham~Sekhar Sahoo, Marianne Arriola, Yair Schiff, Aaron Gokaslan, Edgar Marroquin, Justin~T. Chiu, Alexander Rush, and Volodymyr Kuleshov.
\newblock Simple and effective masked diffusion language models.
\newblock \emph{arXiv preprint arXiv:2406.07524}, 2024.
\newblock URL \url{https://arxiv.org/abs/2406.07524}.

\bibitem[Wang et~al.(2025)Wang, Xu, Jin, Jin, Zhang, and Deng]{wang2025d2f}
Xu~Wang, Chenkai Xu, Yijie Jin, Jiachun Jin, Hao Zhang, and Zhijie Deng.
\newblock Diffusion llms can do faster-than-ar inference via discrete diffusion forcing.
\newblock \emph{arXiv preprint arXiv:2508.09192}, 2025.
\newblock URL \url{https://arxiv.org/abs/2508.09192}.

\bibitem[Wu et~al.(2025)]{wu2025fastdllm}
Chengyue Wu et~al.
\newblock Fast-dllm: Training-free acceleration of diffusion llm by enabling kv cache and parallel decoding.
\newblock \emph{arXiv preprint arXiv:2505.22618}, 2025.
\newblock URL \url{https://arxiv.org/abs/2505.22618}.

\bibitem[Xu et~al.(2025)Xu, Jin, Li, Tu, Long, Tu, Song, Si, Hou, Yan, and Deng]{xu2025lopa}
Chenkai Xu, Yijie Jin, Jiajun Li, Yi~Tu, Guoping Long, Dandan Tu, Mingcong Song, Hongjie Si, Tianqi Hou, Junchi Yan, and Zhijie Deng.
\newblock Lopa: Scaling dllm inference via lookahead parallel decoding.
\newblock \emph{arXiv preprint arXiv:2512.16229}, 2025.
\newblock URL \url{https://arxiv.org/abs/2512.16229}.

\bibitem[Ye et~al.(2025)Ye, Xie, Zheng, Gao, Wu, Jiang, Li, and Kong]{ye2025dream}
Jiacheng Ye, Zhihui Xie, Lin Zheng, Jiahui Gao, Zirui Wu, Xin Jiang, Zhenguo Li, and Lingpeng Kong.
\newblock Dream 7b: Diffusion large language models.
\newblock \emph{arXiv preprint arXiv:2508.15487}, 2025.
\newblock URL \url{https://arxiv.org/abs/2508.15487}.

\end{thebibliography}
